\documentclass[journal]{IEEEtran}
\usepackage{graphicx}
\usepackage{color}
\usepackage[ruled]{algorithm2e}
\usepackage{url}
\usepackage{stfloats}
\usepackage{multirow}
\ifCLASSINFOpdf
\else
\fi
\begin{document}
%
\title{Spirit Distillation: Precise Real-time\\ Semantic Segmentation of Road Scenes\\ with Insufficient Data\\}
%
%
%

\author{Zhiyuan~Wu,
        Yu~Jiang,
        Chupeng~Cui,
        Zongmin~Yang,
        Xinhui~Xue,
        Hong~Qi

\thanks{Manuscript received January ??, ????; revised June ??, ???? and November ?, ????; accepted December ??, ????. Date of publication January ??, ????; date of current version December ??, ????. This work was supported by the National Natural Science Foundation of China under Grant U1234???. The Associate Editor for this paper was ????????.\\ \emph{(Corresponding author: Hong Qi.)}}
\thanks{H. Qi and Y. Jiang are with Key Laboratory of Symbolic Computation and Knowledge Engineering of Ministry of Education, Jilin University, Changchun, China, and also with College of Computer Science and Technology, Jilin University, Changchun, China. (e-mail: qihong@jlu.edu.cn, jiangyu2011@jlu.edu.cn)}
\thanks{Z. Wu, C. Cui, Z. Yang and X. Xue are with College of Computer Science and Technology, Jilin University, Changchun, China. (e-mail: wuzy2118@mails.jlu.edu.cn; cuicp2118@mails.jlu.edu.cn; yangzm2118@mails.jlu.edu.cn; xuexh2118@mails.jlu.edu.cn)}%
}

%
%

\markboth{IEEE TRANSACTIONS ON INTELLIGENT TRANSPORTATION SYSTEMS,~Vol.~xxx, No.~xxx, August~2021}%
{Shell \MakeLowercase{\textit{et al.}}: Bare Demo of IEEEtran.cls for IEEE Journals}
%



\maketitle

\begin{abstract}
    Semantic segmentation of road scenes is one of the key technologies for realizing autonomous driving scene perception, and the effectiveness of deep Convolutional Neural Networks(CNNs) for this task has been demonstrated. 
    State-of-art CNNs for semantic segmentation suffer from excessive computations as well as large-scale training data requirement. Inspired by the ideas of Fine-tuning-based Transfer Learning (FTT) and feature-based knowledge distillation, we propose a new knowledge distillation method for cross-domain knowledge transference and efficient data-insufficient network training, named Spirit Distillation(SD), which allow the student network to mimic the teacher network to extract general features, so that a compact and accurate student network can be trained for real-time semantic segmentation of road scenes.
	Then, in order to further alleviate the trouble of insufficient data and improve the robustness of the student, an Enhanced Spirit Distillation (ESD) method is proposed, which commits to exploit a more comprehensive general features extraction capability by considering images from both the target and the proximity domains as input.
	To our knowledge, this paper is a pioneering work on the application of knowledge distillation to few-shot learning. 
	Persuasive experiments conducted on Cityscapes semantic segmentation with the prior knowledge transferred from COCO2017 and KITTI demonstrate that our methods can train a better student network (mIOU and high-precision accuracy boost by 1.4\% and 8.2\% respectively, with 78.2\% segmentation variance) with only 41.8\% FLOPs (see Fig. 1).
\end{abstract}


%
\IEEEpeerreviewmaketitle

\begin{figure}[!t]
	\centering
	\includegraphics[width=3.2in]{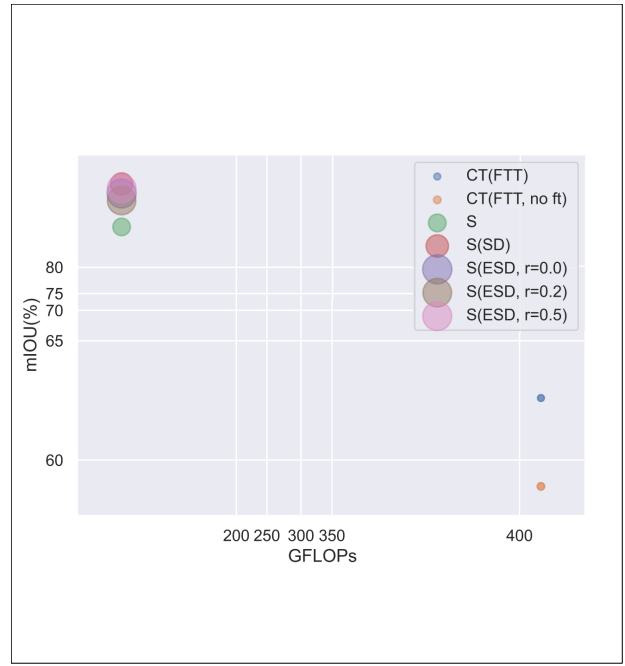}
	\caption{The performance and FLOPs of different architectures with various settings on the first 64 images of Cityscapes \cite{cityscapes}. $CT$ denotes the constructed teacher network, which is the final network gained when the $FFT$ method is adopted. The radius of the bubbles represents high-precision accuracy of respective settings. With $SD$, the student network is able to achieve better segmentation effects with only 41.8\% FLOPs. Further adoption of $ESD$ can improve the robustness of the student network.}
\end{figure}

\section{Introduction}
%
%
%
%


\IEEEPARstart{A}{utonomous} driving technology has made significant progress recently, and is expected to realize a wide range of applications \cite{app1,app2} such as self-driving cars and automated delivery robots. 
As a key technique for self-driving visual analysis, semantic segmentation enables scene perception by recognizing surrounding objects and regions. 
Advances in deep Convolutional Neural Network(CNN) suggests an feasible approach to achieve advanced semantic labeling \cite{ss1,ss2,ss3}, whereas real-time, accurate and robust requirements poses double challenges of limited computational resources and inadequate training data.

Existing state-of-art CNN architectures tend to be time costly for making segmentations, which often fail to meet the real-time demand. For example, DeepLabV3 \cite{dv3} achieves high segmentation accuracy on the Cityscapes \cite{cityscapes} dataset, but the cost of huge FLOPs (with ResNet-50 \cite{cf1} backbone) for a single prediction makes it only able to run less than 19 FPS on an RTX 2060 GPU, even with $\frac{1}{4}$ input resolution. Moreover, driving scenarios are highly variable and contain all kinds of rare situations, e.g., accidents, bad weather, and abnormal driver actions \cite{rare-situations}. Since collecting and labelling sufficient large-scale data is expensive and time-consuming, and it's nearly impossible to cover all kinds of rare cases, the scene perception CNNs tend to be inferior performance and terrible inference on infrequent conditions, which may potentially affect the safety of the self-driving system, hence application prospects are limited.

One instant way to allivate the trouble of inadequate data is to adopt the Fine-tuning-based Transfer Learning($FTT$) \cite{finetune}. 
Based on the assumption that the first few layers of CNNs that converge on huge datasets are able to extract features with universal properties (i.e. general features), e.g., textures and edges, feature analysis layers are required to be constructed and used to replace the rest part of the original network.
Through training under the condition of freezing the weights of the first half and do further fine-tuning, the trained-out network tends to be of great generalization capabilities. 
Nevertheless, since cumbersome backbone networks lead to the huge size of parameters and expensive computational cost, the real-time requirement of autonomous driving is still unable to be met.

To improve the recognition accuracy with insufficient data while reducing the computational effort as much as possible, the idea of transfer few-shot learning and compact model design can be both referenced, which means introducing the prior knowledge of feature representations from CNNs pre-trained on large-scale datasets to compact networks.
As a method with both the capabilities of prior knowledge transfer and model compression, knowledge distillation has great potential for training compact few-shot learning CNNs. 
During the distillation process, bulky and compact networks play the roles of teacher and student, respectively.
However, previous works on knowledge distillation mostly focus on the compression of cumbersome or bagging models.
To the best of our knowledge, no research has yet adopted knowledge distillation to train segmentation models under data insufficiency conditions.

In this paper, we pioneer the application of knowledge distillation to both model compression and few-shot learning, and proposes the Spirit Distillation($SD$). 
Inspired by the ideas of Fine-tuning-based Transfer Learning($FTT$) \cite{finetune} and feature-based knowledge distillation \cite{kd-sum}, we commit to exploit the potential of the student to extract general features discarding the cumbersome backbone. 
Different from previous approaches, $SD$ employs the teacher and student networks that are trained on different datasets, enabling knowledge transference across domains. 
We propose to learn an efficient feature extractor under the supervision of general features extracted by the teacher's activation map generator. 
By prioritizing the learning of stage-generalized feature representation through activation map mimicking, the student network is able to converge better in the subsequent optimization process, and gains a better segmentation performance while satisfying the real-time requirement.

To further alleviate the trouble of insufficient data and improve the robustness of the student, Enhanced Spirit Distillation($ESD$) is introduced, by adjusting the input images in the mimicking phase. 
We proposed to introduce unlabeled images similar to the training dataset as feature extraction materials, ratio-based randomly select and shuffle with the images of the training dataset, which implicitly expanding the data volume and scenarios. The student network that learns from the features of more images possesses better robustness feature extraction capabilities, and achieves a more stable performance after fine-tuning.

In general, our contributions can be summarized as follows:

\begin{itemize}
	\item
	We propound to apply knowledge distillation to both model compression and few-shot learning, and propose the Spirit Distillation($SD$). Through general feature extraction knowledge transference, the student network is able to extract general features with insufficient supervised data, while the time cost of inference meets the real-time requirement.
	\item
	We extend $SD$ to Enhanced Spirit Distillation($ESD$). By introducing the proximity domain to achieve implicit data expansion, the overfitting that the student to little features can be easily avoided, and the robustness of the student is significantly boosted.
	\item
	With the prior knowledge transferred from COCO2017 \cite{coco} and KITTI \cite{kitti}, the performance of our methods have been verified under Cityscapes \cite{cityscapes} benchmark with various settings. Results demonstrate that (1) Spirit Distillation can significantly imporve the segmentation performance of the student network (by 1.8\% mIOU enhancement) without enlarging the parameter size. (2) Enhanced Spirit Distillation can reinforce the robustness of the student(8.2\% high-precision accuracy boosting and 21.8\% segmentation variance reduction) with comparable segmentation results attained.

\end{itemize}

\section{Related Work}
\subsection{Knowledge Distillation}
Knowledge distillation researches on the technical means of training compact student network with the prompt of cumbersome teacher network. It has recently become one of the most popular approaches for model compression. Pioneering works on knowledge distillation mainly focuses on image classification. Hinton \cite{hinton} uncovered the varying similarity between categories. Based on the assumption that the high-entropy teacher network’s output (i.e. soft label) contains richer supervisory information than hard labels (i.e. category label) \cite{soft-and-hard}, a weighted average of soft and hard labels is adopted as supervisory information for student network training. The subsequent FitNet \cite{fitnet} extended the idea of \cite{hinton} during training, by using the hidden layer output of the teacher network to train a thin and deep student network, hoping that the student network learns a transformed intermediate output-based representation.

With the increasingly urgent demand for high-precision real-time CNNs for autonomous driving, subsequent knowledge distillation methods begin to be applied to visual tasks for environment perception, such as object detection and semantic segmentation. Xie et al \cite{xie} normalizes student networks using teacher’s zero and first-order knowledge, and extends the training process to coarsely labelled data. Liu et al \cite{liu1,liu2} proposed the structured knowledge distillation method that considers semantic segmentation as a structured prediction task, measuring the effectiveness by the correlation of both activation maps and segmentation results between teacher and student. He et al. adopt self-encoders \cite{he} to reconstruct intermediate features of the teacher network, and propose affinity modules to capture higher-order dependencies of teacher-student networks’ features.

\subsection{Few-shot Learning}
Few-shot learning provides a solution to the problems in scenarios with insufficient data. Making use of prior knowledge like understanding of the dataset or models trained on other datasets to reduce the dependence of machine learning models on data \cite{few-survey}. Existing few-shot learning methods based on data augmentation \cite{da1,da2,da3}, metric learning \cite{ml1,ml2,ml3}, and initialization \cite{init1,init2,init3} ameliorate the models in terms of supervised empirical growth, hypothesis space reduction, and initial parameter setting, respectively, so as to enhance the generalization ability of the models under the premise of inadequate training data.

\subsection{Compact Model Design}
Compact model design comes up with some specially designed modules to extract and express features. These modules can be directly integrated into the network for training, and a balance can be found between accuracy and efficiency. MobileNetV2 \cite{mobilenetv2} introduces inverted residual with linear bottleneck, and achieve efficient utilization of parameters. Based on this benchmark, MobileNetV3 \cite{mobilenetv3} combines complementary search techniques as well as a novel architecture design into the process of constructing neural networks, which not only reduces FLOPs but also enhances accuracy. GhostNet \cite{ghostnet} further exploits the correlation between activation maps. By using simple linear operations to generate effective activation maps, it has become one of the most advanced compact neural networks by now.

\section{System Model}
We refer to the basic framework of feature-based knowledge distillation \cite{kd-sum}, which introduces both the teacher network($T$) and the student network($S$) in the training procedure. The teacher network adopts state-of-art architecture with pre-trained weights, and the student is compact and efficient.

Carefully designed over-parametric teacher network tend to converge to satisfactory solutions and possesses splendid general feature extraction and representation capacities after training on a large-scale dataset, while the student network is capable of efficient inference, but performs inferiorly when trained directly due to the poor local optimal solution problem\cite{non-conv}.
Hence, there is great potential to improve the performance of the student network through knowledge transference.

The idea of curriculum learning \cite{cur-learning} suggests that prioritizing the learning of simple concepts when learning complex ones tend to be beneficial in improving the final prediction effect of machine learning models.
Inspired by the theory, feature-based distillation enables to transfer the intermediate feature of the teacher, in hinting the optimization of the student network to help it learn teacher's representation.

Fig. 2 displays a general framework of feature-based distillation.
The student is optimized through minimizing the distillation losses($L_D$), which is calculated based on the hidden layer output features of the teacher and the student, through which the student can learn a rich teacher-based intermediate representation.
In most cases, the hidden layer output of the teacher network needs to be reinterpreted(from which we gain $F_T$) to match the size of the student network's intermediate feature($F_S$). So far, we obtain our optimization objective:
\begin{equation}
    W_S^{front} \gets \arg\min {L_D}({F_T},{F_S})
\end{equation}
thereout, we learn the weights of the front part of the student($W_S^{front}$) from its teacher.

Unlike previous methods, the teacher network and the student network in this paper are expected to do segmentation on different domains. Our teacher is pretrained on the source domain($D_s$), and the student is to be trained on the target domain($D_t$). Since there is a huge gap in sample size between the target and the source domains, i.e. and the amount of data of ${D_t}$ is much smaller than that of ${D_s}$, our goal is to improve the performance of $S$ on ${D_t}$, which is insufficient road scenes-oriented semantic segmentation data, to the greatest extent, with powerful knowledge transferred from the representation of $T$ learned from ${D_s}$.

Moreover, the proximity domain($D_p$) that is similar to ${D_t}$ is introduced and adopted as feature extraction material, in providing richer knowledge to enhance the distillation effect. Due to the implicit increment of the amount of data as well as scenario variety, the student network is expected to get richer supervision and performs better.

\begin{figure}[!t]
	\centering
	\includegraphics[width=3.5in]{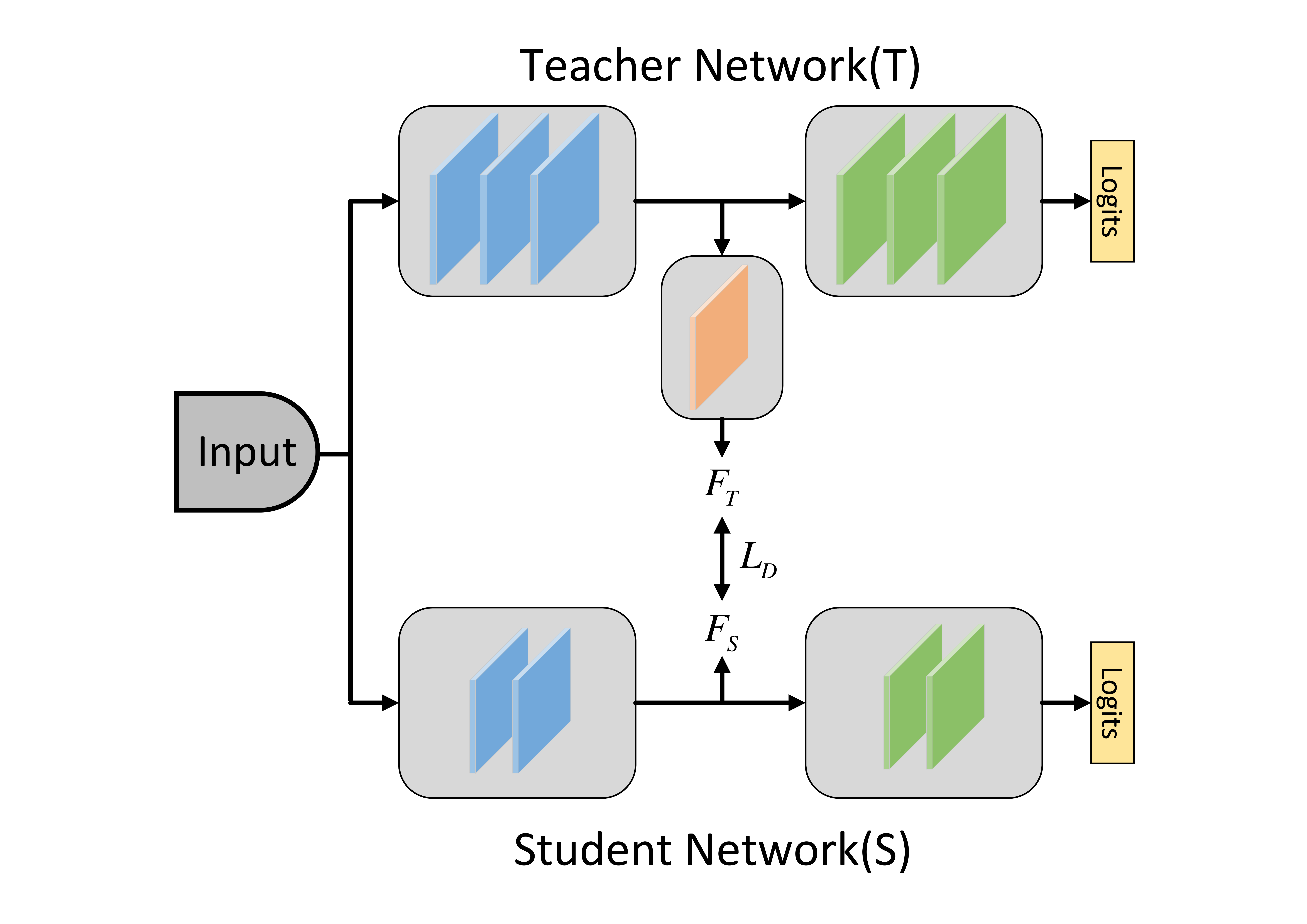}
	\caption{The general framework of feature-based distillation \cite{kd-sum}. The reinterpreted hidden layer output of the teacher supervises the front part of the student for training, through which the knowledge of teacher-based stage-specific feature representation is transferred.}
\end{figure}

\begin{figure*}[!t]
	\centering
	\includegraphics[width=7.2in]{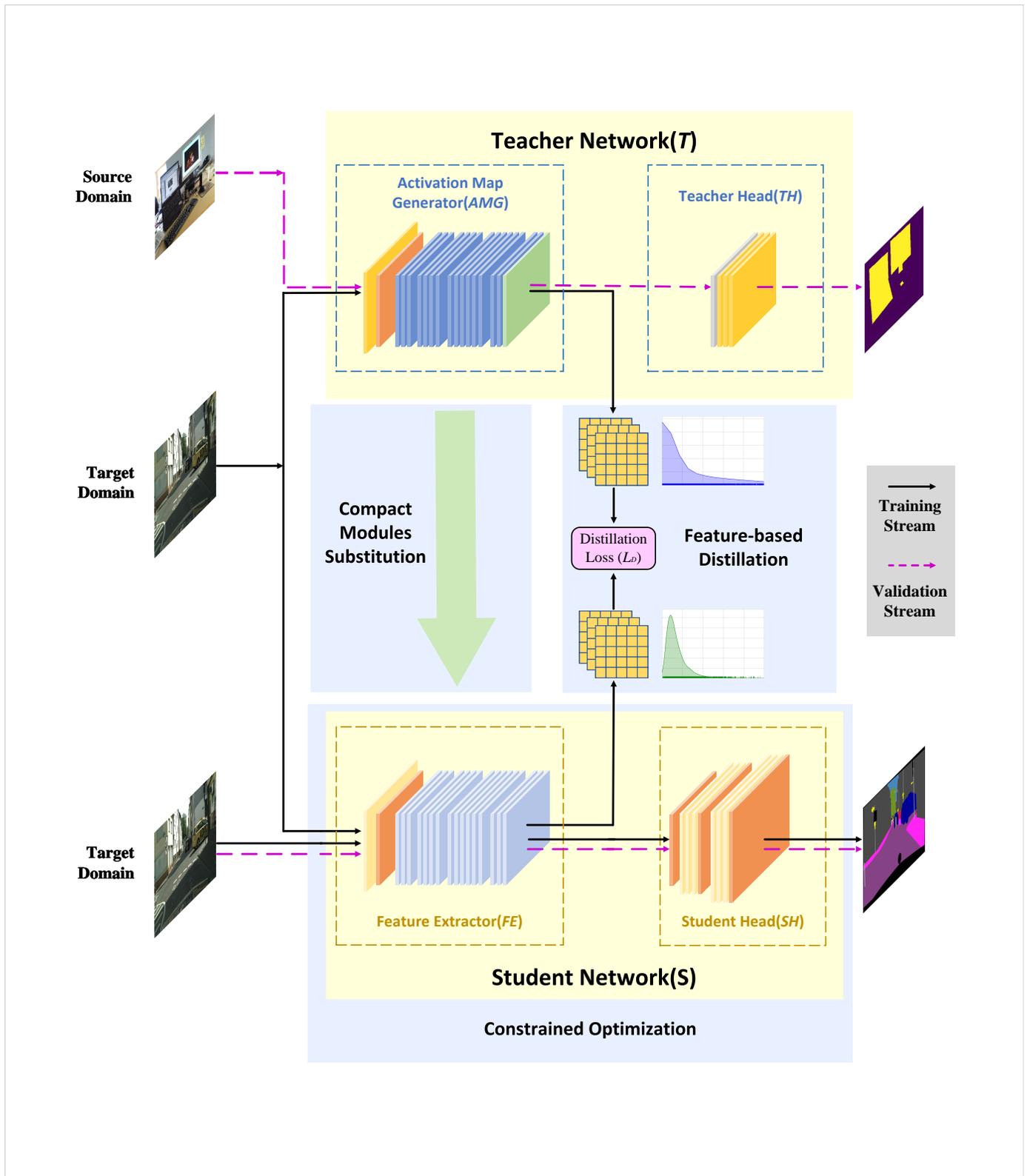}
	\caption{The framework of Spirit Distillation. The teacher network is pre-trained on a large-scale dataset while the student network is used to do segmentation in the domain with insufficient data. The output of the activation map generator guides the feature extractor for training, to achieve the transference of general feature extraction knowledge.}
\end{figure*}

\section{Approach}
In this section, we propose two approaches, Spirit Distillation($SD$) and Enhanced Spirit Distillation($ESD$), to improve the performance of the student network from two perspectives: the transference of the general feature extraction capabilities, and the enriching of the supervised knowledge through data expansion. Our considerations mainly focuses on two aspects: (1) The feature extraction capability of the front part of the large-scale-dataset-pretrained teacher network can be generalized to other datasets, and its feature representations can be effective not only for feature extraction in the target domain, but also for the source domain and other domains of data. (2) The general features of the target domain and proximity domains are in a similar feature manifold, and the introduction of the proximity domain data can prevent the overfitting of the generalized features extracted from the target domain, while also implicitly providing richer data information.

In the following parts, we will introduce both of our two methods in detail, and displays the formal description of Enhanced Spirit Distillation.

\subsection{Spirit Distillation}
The general features of the teacher network pretrained on a large-scale dataset possess universal properties and generalization ability. But adopting $FTT$ method that directly utilize the first few layers of the teacher network for feature extraction will result in the final CNN containing computational costly backbone, which is difficult to meet the real-time demand. 
To address this problem, we fully exploit the potential of the compact student network in attaining similar output as its teacher with much fewer FLOPs, and propose the Spirit Distillation($SD$).

We display the framework of Spirit Distillation in Fig. 3. As shown, $SD$ is different from previous methods[???], in the aspect of employs the teacher and the student networks that trained on different datasets, enabling knowledge transference across domains.
Before training, both of the two networks are divided into feature extraction part and representation head, and the feature extraction part of the two are denoted as the activation map generator($AMG$) and the feature extractor($FE$), for generating supervised features and $S$'s efficient general feature extraction, respectively.
$TH$ / $SH$ parses the output activation maps of the first part of their respective networks, and obtains segmentation results of their respective domains. 
We input the images of $D_t$ to both of the two networks, and the features of them are obtained from $AMG$.
Then, the weights of $FE$ would be optimized based on the distillation loss ($L_D$) between $AMG$'s and $FE$'s output, to achieve the transference of the general feature extraction representation.
The last step is to conduct further optimization to realize accurate segmentation.

The next subsections will introduce the detailed process of Spirit Distillation.

\subsubsection{\textbf{Teacher Network and Student Network}}
Given a bulky pretrained teacher network $T$, we divide it into two parts referencing to the deviation between ${D_s}$ and ${D_t}$. The larger the gap between the two domains, the closer the division is to the front of the teacher network. 
In this way, we gain the activation map generator($AMG$) which is the first part, and the teacher head($TH$) that to be the second one.
Obtaining the $S$'s feature extractor ($FE$) by replacing the convolutional layers of $AMG$ with compact modules(e.g. group convolution\cite{alex}), to prepare for the ground for efficient feature extraction. Constructing $S$ by designing the efficient feature analysis part (i.e., student head, denoted as $SH$) for semantic segmentation of road scenes, and $SH$ is stacked after $FE$. As such, the inference cost of $S$ is much cheaper than that of $T$, and its $FE$ has the potential to extract general features just like $AMG$, with even stronger generalization capability due to the smaller parameter size.

\subsubsection{\textbf{Feature-based Distillation}}
We input images of ${D_t}$ into $AMG$, and can gain their general features(i.e., the output of the $AMG$, denoted as ${y^{AMG}}$). 
Suppose that the general features extracted by $AMG$ are "spirits" of the $T$'s general feature extraction representation. 
These general features are less relevant to a specific domain and a particular network architecture compared with the hidden layer output of bulky networks converged on only the target $D_t$'s training data. 
The rich semantic information of "spirit" for supervision is helpful to guide $FE$ to optimize toward extracting effective features for $D_t$.
As a result, we take ${y^{AMG}}$ as the optimization objective of the feature extractor ($FE$, whose output is denoted as ${y^{FE}}$), and transfer the "spirit" by minimizing the distillation loss ($L_D$), which is defined as mean square error loss, as shown in equation (2).
\begin{equation}
{L_D} = \frac{1}{{|P^{AMG}|}}\sum\limits_{pos \in P} {\sum\limits_{i = 1}^n {{{(y_{pos,i}^{AMG} - y_{pos,i}^{AMG})}^2}} }
\end{equation}
where $P^{AMG}$ is the set of positions of pixels in $y^{AMG}$, $n$ is the number of classes, and $y_{pos,i}^{AMG}$ is the gray value corresponding to the $pos$ position with channel $i$ of $y^{AMG}$.

\subsubsection{\textbf{Constrained Optimization}}
Having transferred the knowledge from the teacher network, further optimization is required for precise segmentation. This paper adopts the similar optimization strategy as fine-tuning-based transfer learning($FTT$), to train $S$ with a frozen $FE$, followed by a small learning rate optimization for the overall weights. We adopt pixel-average cross-entropy as our training loss function(${L_{\rm{S}}}$), as shown in equation (3):
\begin{equation}
{L_{\rm{S}}} = {\rm{ - }}\frac{1}{{|P^{pred}|}}\sum\limits_{pos \in P} {\sum\limits_{i = 1}^n {y_{pos,i}^{label}} } \log (y_{pos,i}^{pred})
\end{equation}
where $y^{pred}$ means the output of the student network, $P^{pred}$ is the set of positions of pixels in $y^{pred}$, and $y^{label}$ means the n-channel feature vector based on the category expansion of the n-category segmentation ground truth.

\begin{table*}[]
\centering
\begin{tabular}{lccll}
\hline
\multicolumn{1}{c}{Dataset} & \multicolumn{1}{l}{Num of Labeled Images} & Resolution     & \multicolumn{1}{c}{Preprocess}                                                                     & \multicolumn{1}{c}{Usage}                                                                                                                                                                                              \\ \hline
COCO2017                    & $100K+$                                     & -              & \multicolumn{1}{c}{-}                                                                              & \begin{tabular}[c]{@{}l@{}}The subset contains the same class as Pascal VOC is used\\ to pre-train the teacher network.\end{tabular}                                                                                   \\ \hline
Cityscapes                  & $3475$                                      & $2048 \times 1024$      & \begin{tabular}[c]{@{}l@{}}random crop, random horizonal \\ flip, max pooling\end{tabular}          & \begin{tabular}[c]{@{}l@{}}Only the first 64 images of Aachen is chosen for feature-\\-based distillation and constrained optimization.\end{tabular}                                                                   \\ \hline
KITTI                       & $15K$                                       & about $1224 \times 370$ & \begin{tabular}[c]{@{}l@{}}resize, random crop, random \\ horizonal flip, max pooling\end{tabular} & \begin{tabular}[c]{@{}l@{}}When adopting the modified Spirit Distillation, its images \\ are randomly shuffled with Cityscapes ones in the feature-\\-based  distillation process, as general feature extraction\\ materials.\end{tabular} \\ \hline
\end{tabular}
\setlength{\abovecaptionskip}{15pt}%
\setlength{\belowcaptionskip}{10pt}%
\caption{The main properties, preprocessing methods and usage of the datasets adopted in this paper. We transfer the knowledge from COCO2017 \cite{coco} and KITTI \cite{kitti}, for hinting the training of student networks on Cityscapes-64 \cite{cityscapes}}
\end{table*}

\subsection{Enhanced Spirit Distillation}
With insufficient samples, the extraction of general features from the front part of the model is a sufficient condition for it to achieve generalizable segmentation results.
Adopting the Spirit Distillation method, we extract features of $D_t$ images through $AMG$, utilizing which for $FE$’s training.
During the feature-based distillation process, the knowledge of noise removal and image reinterpretation learned from ${D_s}$ is implicitly transferred.

However, since the $D_t$ dataset is largely undersampled from real scenes, the required diversity general features cannot be fully obtained by simply reinterpreting the images from the ${D_t}$ dataset.
Consequently, the fitting of $FE$ to the features extracted from $D_t$ images can be regarded as another form of overfitting, i.e., it fits the function values of $D_t$ images, and the function mapping is realized by the convolution operations of $AMG$.

In fact, the meta-knowledge reflected by $AMG$ tend to be more abstract than how to extract general features of such images from both $D_s$ and $D_t$. It is expected that the general feature extraction approach of ${D_t}$ can be appropriately generalized to datasets in domains similar to ${D_t}$.
On the contrary, learning features of a similar domain (proximity domain, denoted as $D_p$) provide references to the learning of ${D_t}$'s feature extraction approach, which may lead $FE$ to mimic out the more abstract and applicable parts of the how $AMG$ extract general features of domains like ${D_t}$ and ${D_p}$.
Moreover, the proximity domain may implicitly transfer a richer information of scenarios.
By the reason of the $D_p$ dataset represents a relatively dense sampling of a feature space similar to $D_t$, even without segmentation labels, adopting $D_t$'s images as feature extraction materials and gains supervisory features would also give a reasonable objective for $FE$'s imitation, and can compensate for the problem of insufficient data on ${D_t}$.
Based on the assumptions above, we extend $SD$ in terms of data inputting, by shuffling ${D_t}$ and $D_p$'s images together, extracting their features, and allowing the student network to imitate. 
This method expects to be executed in substitution with the input of images of $D_t$ during the feature-based distillation process, as shown in Fig. 4, and we names the newly integrated transferring and training scheme Enhanced Spirit Distillation($ESD$).

\begin{figure}[!t]
	\centering
	\includegraphics[width=3.2in]{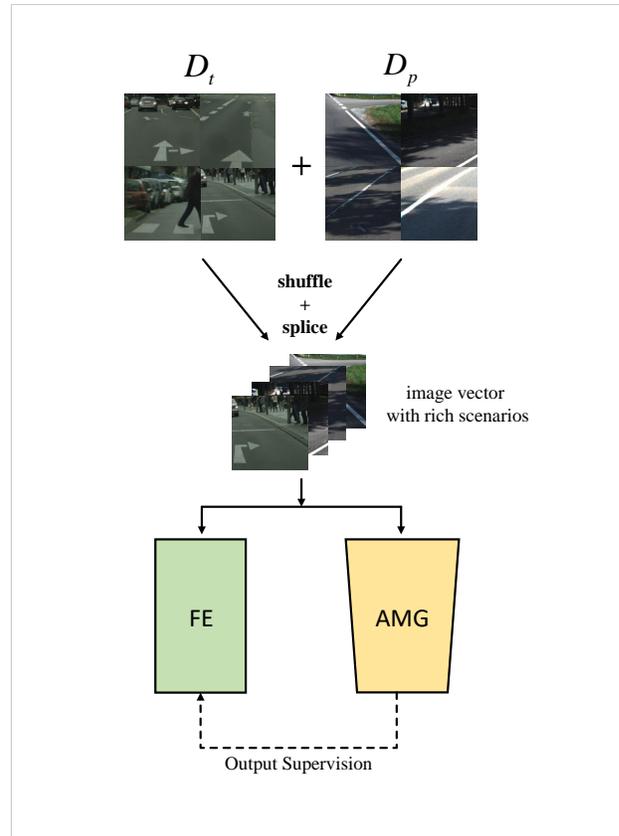}
	\caption{The alternative strategy for input images in $ESD$. Introduce the images of the proximity domain, and shuffle them with target domain images as feature extraction material, to provide richer knowledge. The general features of proximity domain contribute to a more complete sampling of the general features manifold of the target domain, while providing a richer data scenario, e.g., the representation of shadow segmentation can be learned from the proximity domain.}
\end{figure}

Comparing to the supervision of the $FE$ using only the general features of the ${D_t}$ images, introducing ${D_p}$ for feature extraction, has the advantages as follows:

Since $T$ is trained on a large-scale dataset, the mapping accomplished by its $AMG$ is a valid effective mapping scheme for various domain datasets to their general features.
It would be the best if $FE$ is able to accomplish a constant mapping to $AMG$, whereas the smaller capacity makes the idea unable to realize.
The use of ${D_t}$ data as feature extraction materials for $AMG$ and supervising the training of $FE$ with its feature can lead to an effective imitation of general features of ${D_t}$; 
according to the theory of curriculum learning \cite{cur-learning}, this will effectively enhance the final result of the $S$. 
However, due to the inadequacy of data in ${D_t}$, the $FE$ tends to learn its extraction approach based only on features sparsely sampled from ${D_t}$'s general feature space.
The sparsity of feature sampling leads to the incompleteness of the representation of the pattern in the feature space where it is located, and thus the imitation of $S$ tends to be more affected by noise and suffers from sharp inferred fields.
And since ${D_p}$ and ${D_t}$ are sampled in similar scenarios, the feature space of their general features is often close to that of ${D_t}$, hence shuffling both of their data and utilizing them for feature extraction in supervision can explore the representation of $T$'s mapping from more perspectives, which plays a role in enriching the knowledge, and the imitation of student networks will be more adequate and stable.

In addition, the images in ${D_p}$ dataset cover a wider range of scenarios that are not available from ${D_t}$, and such knowledge can be learned to benefit the generalization and the robustness of $S$.
For instance, for the two road scenes segmentation datasets in $D_p$ and $D_t$, neither of the images in the $D_t$ dataset contains shadows, while some of the images in $D_p$ contains(see APPENDIX B).
During the feature-based distillation process, the student network will not mimic out the features extracted from images with shadows, hence the approach of how to extract shadows' features are hard to be learned, and thus the segmentation result of these parts tend to be poor. 
However, if there are shadows in the images of $D_p$, the addition of these kinds of images during the distillation process will add a new feature extraction perspective to the student network when imitating, i.e., how the features of the images with shadows ought to be extracted, and thus the student network learns a more comprehensive general feature extraction approach, which may be able to cope with even some scenarios with the absence of $D_t$, like shadows, and will perform more robustness in rare cases.

\subsection{Formal Description of Enhanced Spirit Distillation}
This section provides a formal description of the overall procedure of Enhanced Spirit Distillation. The algorithm takes the weights of pre-trained teacher network ${W^T}$(whose weights of $AMG$ part corresponds to ${W^{AMG}}$), the weights of randomly initialized student network ${W^S}$ (whose weights of $FE$ and $SH$ parts correspond to ${W^{FE}}$ and ${W^{ST}}$), the distillation loss $L_D$, the inference loss $L_S$, the target domain dataset ${D_t}$, the proximity domain dataset ${D_p}$, and the optimizer $optimizer_i$ of the $i^{th}$ stage of training(with hyper settings) as inputs, and takes trained ${W^S}$ as output. Define ${W_1},{W_2},...{W_k}$ as the weights of layers $\{ {W_1},{W_2},...{W_k}\}$, ${\{ {W_1},{W_2},...{W_k}\} ^X}$ as the output of data $X$ after the transformation operation of each layer, and $W_k^*$ as the result of a certain iteration update of ${W_k}$.

\begin{algorithm}
	\caption{Enhanced Spirit Distillation}
	\KwIn{${W^T}$(${W^{AMG}}$), ${W^S}$(${W^{FE}}$, ${W^{ST}}$), $L_D$, $L_S$, ${D_t}$, ${D_p}$, $optimize{r_i}$}
	\KwOut{Trained ${W^S}$}
	\textbf{Stage 1: Feature-based Distillation} 

	\While{$FE$ not convergence}{
		${X_1}\gets shuffle\_select({D_p},{D_t})$\;
		${W^{FE*}}\gets {W^{FE}} - optimize{r_1}(\nabla {L_D}({\{ {W^{AMG}}\} ^{{X_1}}},{\{ {W^{FE}}\} ^{{X_2}}}))$\;
	}
	\textbf{Stage 2: Frozen Training}

	\While{$SH$ not convergence}{
		${X_2}\gets random\_choice({D_t})$\;
		${W^{SH*}} \gets {W^{SH}} - optimize{r_2}(\nabla {L_S}({\{ {W^T}\} ^{{X_2}}},{\{ {W^S}\} ^{{X_2}}}))$\;
	}
	\textbf{Stage 3: Fine-tuning}

	\While{$S$ not convergence}{
		${X_3}\gets random\_choice({D_t})$\;
		${W^{S*}}\gets {W^S} - optimize{r_3}(\nabla {L_{\rm{S}}}({\{ {W^T}\} ^{{X_3}}},{\{ {W^S}\} ^{{X_3}}}))$\;
	}
\end{algorithm}

\section{Experiments}

\subsection{Datasets}
We introduce the datasets adopted in our experiment, and their preprocessing methods and usages are displayed in Table I.
\subsubsection{COCO2017}
COCO \cite{coco} is a large-scale dataset with labels covering a variety of recognition tasks, including segmentation. COCO2017 is the version of the dataset updated in 2017 with 100K+ labeled images. In our experiment, the teacher network has been pre-trained on a subset of COCO train2017, whose weights are at \textcolor[rgb]{1,0,0}{$download.pytorch.org/models/deeplabv3\_resnet50\_coco-cd0a2569.pth$}. The dataset contains common objects in autonomous driving scenarios, such as cars, motorcycles, and people.

\subsubsection{Cityscapes}
Cityscapes \cite{cityscapes} is an urban road scenes-oriented dataset for semantic segmentation research. The images in the dataset were collected from 50 different cities, of which 3475 were given fine annotations, and were pre-split into 2975/500 images for training and validation, respectively. To create the condition of inadequate samples, we only choose the first 64 of the training images of Aachen for training(denoted as Cityscapes-64), and test the segmentation effect through the validation set.

\subsubsection{KITTI}
As one of the largest datasets for evaluating computer vision algorithms in autonomous driving scenarios, KITTI \cite{kitti} covers a variety of evaluation tasks such as stereo, tracking, and detection. In this paper, we adopt the training images for 2d object detection as the image source for feature extraction in Enhanced Spirit Distillation.

\subsection{Network Architecture}
We adopt DeepLabV3\cite{dv3}(resnet-50\cite{cf1} backbone, pre-trained on COCO2017) as the teacher network. To construct the feature extractor, all of the convolutional layers of its backbone are replaced by group convolutions\cite{alex}, each group of which being the greatest common factor of the number of input and output channels. The student head is constructed by replacing the ASPP and subsequent layers with a SegNet-like \cite{segnet} decoder structure, i.e., two groups of $3 \times (Conv+BN+ReLU)$ stack with bilinear up-sampling modules to achieve resolution increment and pixel-level classification. The convolution layers of the decoder also adopt group convolutions in the same setup as the $FE$'s construction. Binary classification is performed on the Cityscapes-64 dataset to distinguish between roads and backgrounds. In APPENDIX A, we display the detailed architecture of the student networks, shown in TABLE IV and TABLE V.

\begin{table*}[]
\centering
\begin{tabular}{lcccc}
\hline
\multicolumn{1}{c}{Method}                      & \multicolumn{1}{l}{GFLOPs} & \multicolumn{1}{l}{Param(M)} & \multicolumn{1}{l}{mIOU(\%)} & \multicolumn{1}{l}{HP-Acc(\%)} \\ \hline
CT(FFT \cite{finetune}, without fine-tuning) & 405.7                      & 23.6                         & 62.6                         & 1.4                            \\ \hline
CT(FFT \cite{finetune}, with fine-tuning)    & 405.7                      & 23.6                         & 58.9                         & 2.6                            \\ \hline
S                           & \textbf{169.4}                      & \textbf{9.5}                          & 81.7                         & 81.2                           \\ \hline
Ours: S(SD)                        & \textbf{169.4}                      & \textbf{9.5}                          & \textbf{83.5}                         & \textbf{84.6}                           \\ \hline
\end{tabular}
\setlength{\abovecaptionskip}{15pt}%
\setlength{\belowcaptionskip}{10pt}%
\caption{The performance on the Cityscapes dataset in comparison with normal training and FTT. All the networks are trained only on the first 64 images of Cityscapes.}
\end{table*}

\begin{table}[]
\centering
\begin{tabular}{clccc}
\hline
\multicolumn{2}{c}{Method}                                 & mIOU(\%)      & HP-Acc(\%)    & Var($10^{-3}$)     \\ \hline
\multicolumn{2}{c}{S}                                      & 81.7          & 81.2          & 5.77          \\ \hline
\multicolumn{2}{c}{Ours: S(SD)}                             & \textbf{83.5} & 84.6          & 5.70          \\ \hline
\multicolumn{1}{l}{} & r=10.0 & 82.2          & 85.4          & 5.12          \\ \cline{2-5} 
\multicolumn{1}{l}{}                             & r=5.0  & 81.9          & 85.6          & 5.20          \\ \cline{2-5} 
\multicolumn{1}{l}{}                             & r=3.0  & 82.2          & 84.6          & 5.21          \\ \cline{2-5} 
\multicolumn{1}{l}{Ours: S(ESD)}                             & r=1.0  & 82.3          & 84.2          & \textbf{4.52} \\ \cline{2-5} 
\multicolumn{1}{l}{}                             & r=0.5  & \textbf{83.3} & \textbf{89.2} & 4.71          \\ \cline{2-5} 
\multicolumn{1}{l}{}                             & r=0.2  & 82.8          & \textbf{89.0} & \textbf{4.48} \\ \cline{2-5} 
\multicolumn{1}{l}{}                             & r=0    & \textbf{83.1} & \textbf{89.4} & \textbf{4.51} \\ \hline
\end{tabular}
\setlength{\abovecaptionskip}{15pt}%
\setlength{\belowcaptionskip}{10pt}%
\caption{Comparison results on normal training, spirit distillation, and enhanced spirit distillation with different $r$ values. Enhanced spirit distillation can effectively improve the robustness of the student network with comparable segmentation effect.}
\end{table}

\subsection{Implement Details}
\subsubsection{Metrics}
We adopt mean intersection over union($mIOU$) as the index to measure the segmentation effect. Its definition is given by:
\begin{equation}
mIOU_i = \frac{1}{{k + 1}}\sum\limits_{j = 0}^k {\frac{{TP_{j,k}}}{{FN_{j,k} + FP_{j,k} + TP_{j,k}}}} 
\end{equation}

\begin{equation}
mIOU = \frac{1}{n}\sum\limits_{i = 1}^n {mIO{U_i}} 
\end{equation}
where $k$ is the number of non-background categories, $n$ is the number of images in the validation set, and $TP_{j,k}$, $FN_{j,k}$, $FP_{j,k}$ are the number of pixels in the given $j^{th}$ image with $k^{th}$ category classification results of true positive, false negative, and false positive, respectively.

The variance of $mIOU$($Var$) and high-precision segmentation accuracy($HP \rule[2.95pt]{0.1cm}{0.05em} Acc$) in evaluating the robustness of the network, as defined in equation 5 and 6.

\begin{equation}
Var = \frac{1}{n}\sum\limits_{{\rm{i}} = 1}^n {{{(mIO{U_i} - mIOU)}^2}} 
\end{equation}

\begin{equation}
HP \rule[2.95pt]{0.1cm}{0.05em} Acc = \frac{1}{n}\sum\limits_{i = 1}^n {\delta ({mIOU_i})} 
\end{equation}
where $\delta$ is an truncation function,

\begin{equation}
\delta (x) = \left\{ \begin{array}{l}
1,x > \theta \\
0,otherwise
\end{array} \right.
\end{equation}

Here, $\theta$ is the high-precision segmentation decision threshold, and an image is considered to be correctly segmented when its $mIOU$ is higher than $\theta$. In this paper, we take $\theta=75\%$ as the decision threshold for high-precision segmentation.

Furthermore, the size of model parameters and floating point operations are adopted in respectively measuring the compactness and inference efficiency of the network.

\subsubsection{Hyper-parameter Settings}
We take a comparison experiment on whether or not to adopt the Spirit Distillation($SD$) method. We also employ the $FTT$ on the network that stacks $AMG$ and the $SH$ (denoted as constructed teacher network($CT$)), and the former part is frozen in the first training stage. We directly train the student network using stochastic gradient descent($SGD$) with momentum $0.9$ and learning rate ${10^{ - 2}}$. For distillation process, a learning rate of $3 \times {10^{ - 3}}$ and a momentum of $0.99$ are used until convergence. Constrained optimization requires freezing weights of $FE$. The training of the remaining portion adopts a momentum of $0.9$ with a learning rate of ${10^{ - 2}}$. Fine-tuning sets the learning rate of the entire network to $5 \times {10^{ - 5}}$ and the momentum to $0.99$. l2 weight penalty is adopted in all cases, with a decay constant of $3 \times {10^{ - 3}}$. Moreover, a data enhancement scheme with random cropping ($512 \times 512$) and random horizontal flipping is adopted, with max pooling(kernel\_size=2) adopted before 
input. 
For the validation of the Enhanced Spirit Distillation($ESD$), we set up a series of different scales to control the ratio $r$ of the input ${D_t}$ data to the ${D_p}$ data during distillation, and conducted the experiments separately. The ${D_t}$ images are preprocessed in the same way as that adopted for $SD$ except that they were previously resized before preprocessing. The preprocessing schemes for all images are shown in Table I.

\begin{figure}[!t]
	\centering
	\includegraphics[width=3.2in]{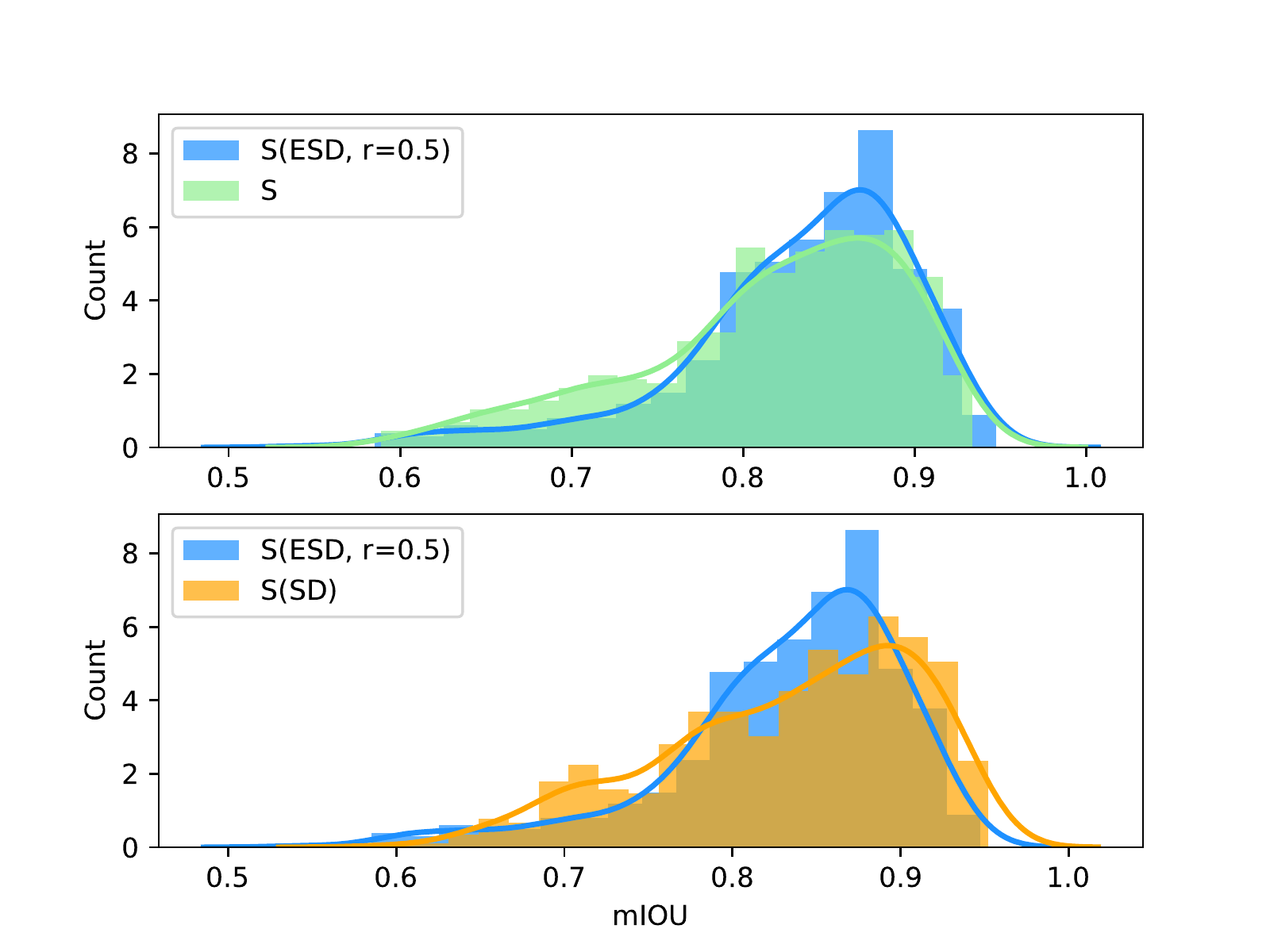}
	\caption{Comparison of the distribution of segmentation results of the three training methods on Cityscapes \cite{cityscapes} validation set. Our $ESD$ can effectively improve the average segmentation effect of the student network, while significantly improving the high-precision segmentation accuracy with better robustness of the final student network, on the condition that $mIOU$ comparable to that adopting only the $SD$ method.}
\end{figure}

\begin{figure*}[!t]
	\centering
	\includegraphics[width=7.2in]{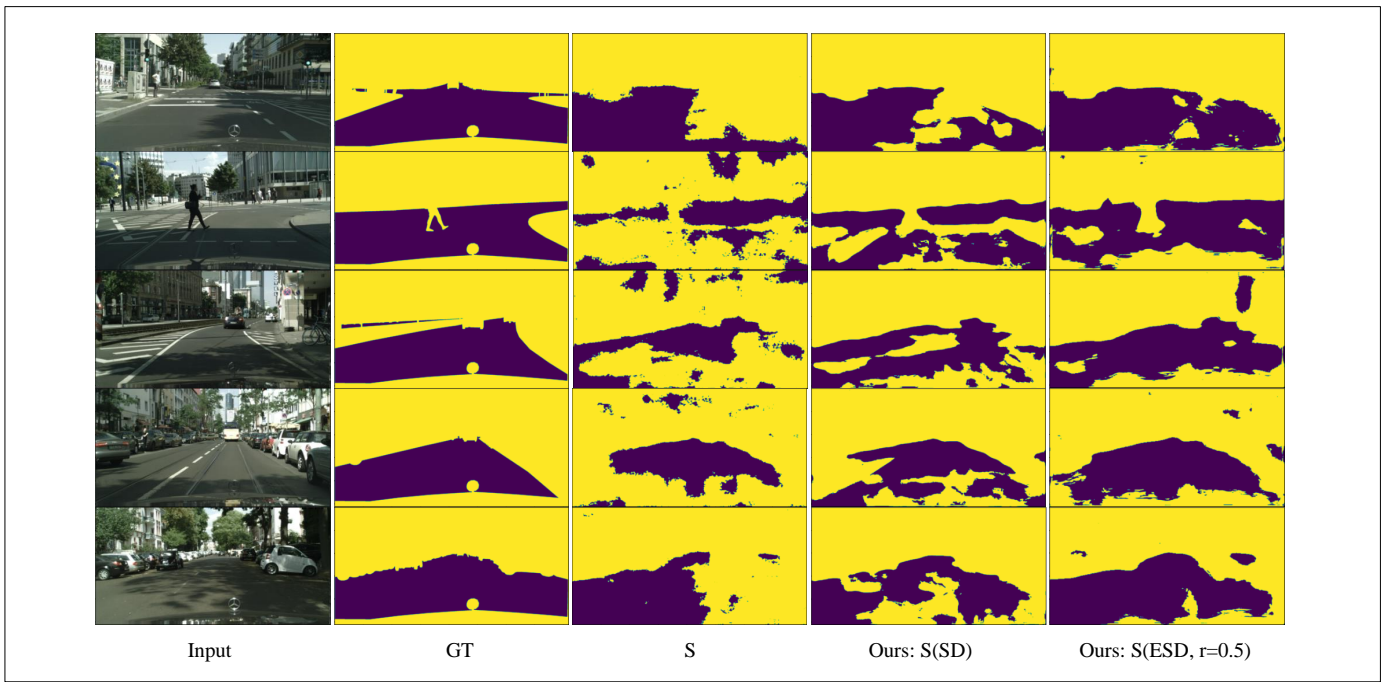}
	\caption{Comparison of segmentation results on Cityscapes-64. (1) Input images. (2) Ground truth. (3) The segmentation results of the student network. (4)The segmentation results of the student network that adopts Spirit Distillation. (5) The result of the student network that adopts Enhanced Spirit Distillation, with $r$ value set to be 0.5.}
\end{figure*}

\begin{figure}[!t]
	\centering
	\includegraphics[width=3.2in]{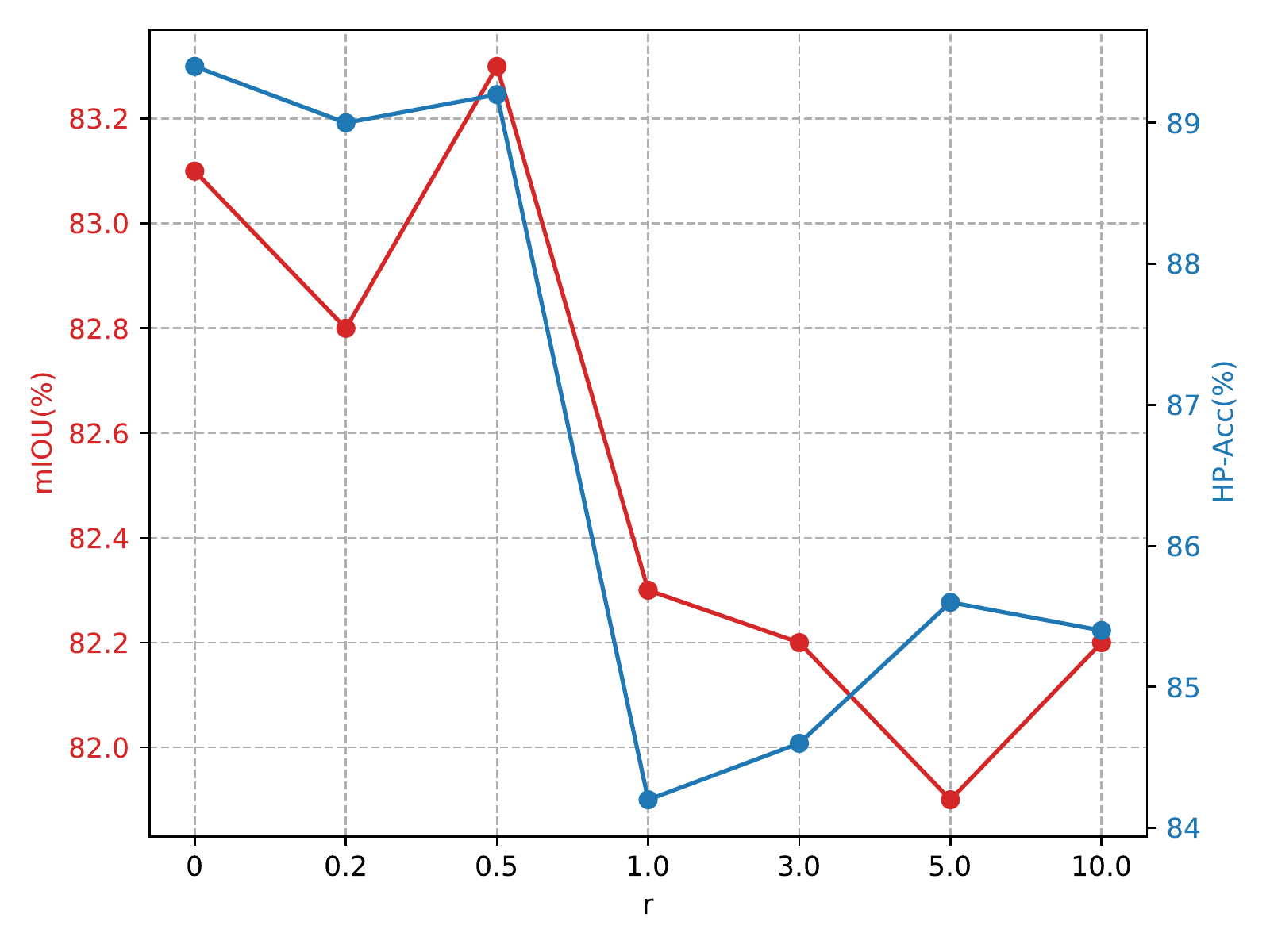}
	\caption{Variation of $HP \rule[2.95pt]{0.1cm}{0.05em} Acc$ and $mIOU$ across different $r$ values. Using smaller $r$ values by default, and appropriately increase the proportion of proximity domain data will significantly help improve the segmentation effect as well as the robustness.}
\end{figure}

\begin{figure}[!t]
	\centering
	\includegraphics[width=3.5in]{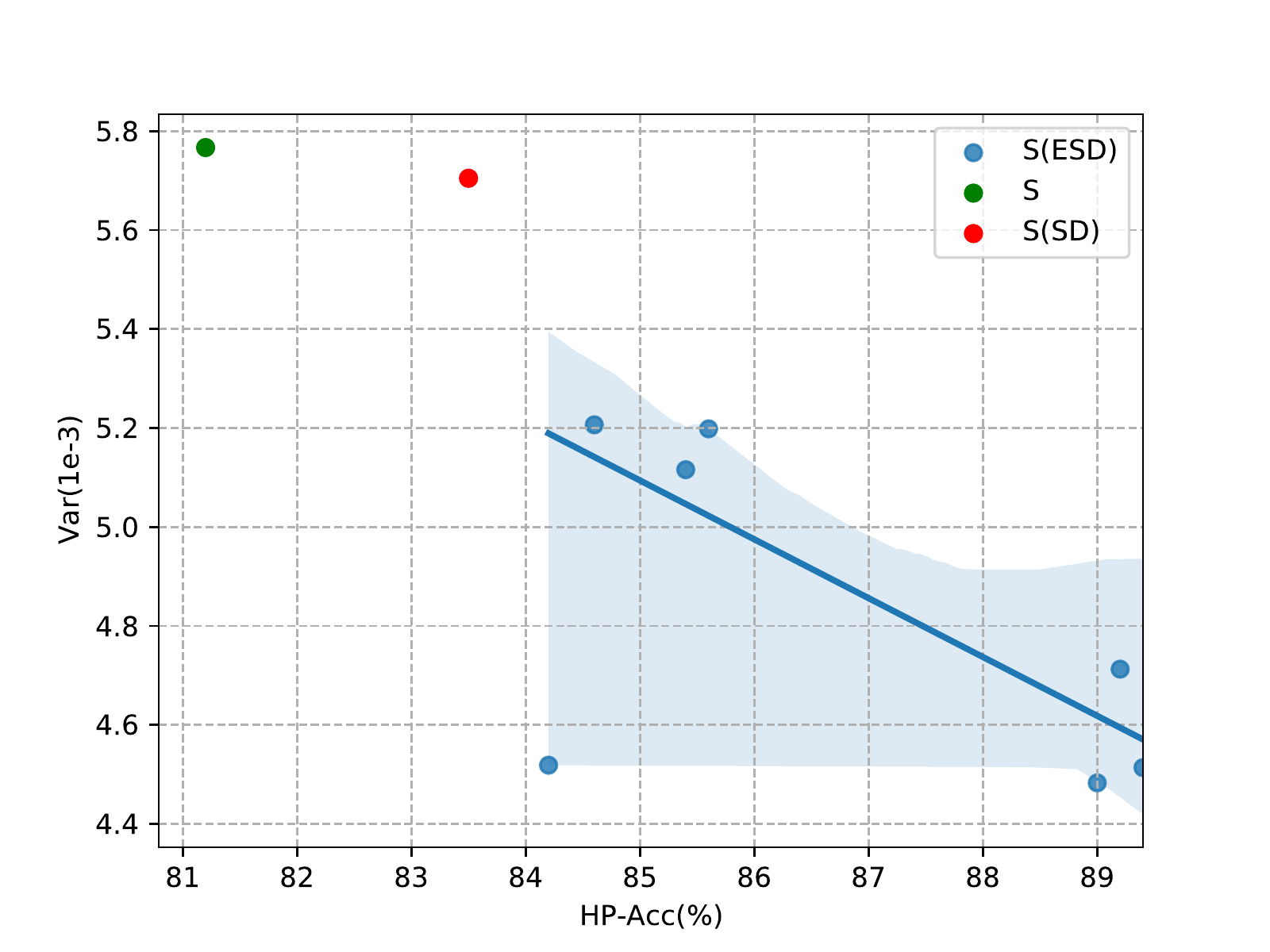}
	\caption{$HP \rule[2.95pt]{0.1cm}{0.05em} Acc$ with variance results of the student network adopting normal training scheme, Spirit Distillation, Enhanced Spirit Distillation with various settings. Enhanced Spirit Distillation has a significant effect on improving the robustness of the student network under various hyper-settings.}
\end{figure}

\subsection{Results}
TABLE II displays the segmentation results of the student network($S$) trained with Spirit Distillation($SD$) along with the results of that trained follows normal training scheme and Fine-tuning-based Transfer Learning($FTT$).
In the validation experiments of Enhanced Spirit Distillation($ESD$), we respectively train the student network under different $r$ settings, and obtain the results shown in Table III.
For result perception, we calculate the segmentation results measured by $mIOU$ on the validation set under three training settings, and formulate a histogram in displaying the distribution of the segmentation results, as shown in Figure 5.
For further result analysis, we display the variation of high-precision segmentation accuracy($HP \rule[2.95pt]{0.1cm}{0.05em} Acc$) across different $r$ values, and the $HP \rule[2.95pt]{0.1cm}{0.05em} Acc$ with variance results in a variety of settings, as shown in Fig. 7 and Fig. 8. 
We also visualize the segmentation results of some image samples that are segmented by the student networks trained in three different methods, and the results are shown in Figure 6.

The results in TABLE II show that the student network using $SD$ outperforms the normally trained $S$ as well as the fine-tune-transferred $CT$. In addition, with inference efficiency significantly improved, $SD$ also prevents the final network from over-fitting due to the large network size as well as under-fitting for the sake of freezing weights.

Fig. 5 compares the distribution of segmentation results of the normally trained $S$ with the results of the $S$s that are trained using the $SD$ method and the $ESD$.
It can be seen that the segmentation effect of the student network is improved with either $SD$ or $ESD$ adopted. Relatively, the student network trained with $ESD$ can perform splendid segmentation effects in more cases, and the proportion of very poor results is significantly reduced. Hence, the $ESD$ method can largely improve the robustness of the student network.

TABLE III and Fig. 7 display the segmentation results of $ESD$ on different $r$ settings. When the value of $r$ is set small, i.e. the target domain accounts for a smaller proportion of the images used for feature extraction, the final obtained $S$s tend to gain a higher high-precision accuracy as well as $mIOU$, which indicates the effectiveness of the introduction of the proximity domain data. We can also conclude that using small $r$ values by default, which means appropriately increase the proportion of proximity domain data will help improve the segmentation effect as well as the robustness.

Fig. 8 reveals the relationship between high-precision accuracy and the variance of segmentation results when adopting different training settings.
From which we discover that the $ESD$ method effectively improves the high-precision accuracy while keeping the variances small numbers. The comprehensive learning of the general features extracted from $T$ helps prevent unstable output and enhances robustness.

The visualization of the segmentation results in Fig. 6 confirms the validity of our method in knowledge transfer and enhancement. It can be observed that the segmentation results of the undistilled $S$ are rather unsatisfactory for the shadow parts of the images. 
After adopting the $SD$ method, the new $S$ has improved the segmentation results of these parts, which is able to distinguish the road scene from the partial shadows. 
Adopting the $ESD$ method on top of this, the $S$ would capture the global representation for shadow segmentation more completely, and can distinguish the road part with shadows of the images more as a whole. 
The segmentation of the sample images further confirms the usefulness of our method for imporving the segmentation effectiveness as well as the robustness under complex scenarios.







\section{Conclusions and Future Works}
In this paper, we propose the Spirit Distillation($SD$) to allow the student network to mimic the teacher network to extract general features, through transferring the knowledge of general features extracted by the teacher network in the domain with insufficient data. In this way, the potential of the student to extract general features is well exploited.
We also extend $SD$ to Enhanced Spirit Distillation($ESD$) to further boost the robustness of the student network through introducing images of the proximity domain as feature extraction materials during feature-based distillation process.
Experiments demonstrate that our methods can effectively achieve cross-domain knowledge transference, and significantly boost the segmentation effect as well as the robustness of compact models even with insufficient data.

In addition, this paper pioneers the application of knowledge distillation to both model compression and few-shot learning, setting the precedent for efficient, low-data-dependent model training. 
Future works will include extending our approach to other visual applications, as well as conducting domain transformation of feature extraction materials using approaches like conditional generative adversarial network training.

 \appendices
  \section{STUDENT NETWORK INSTANTIATION}
 In this paper, we construct the feature extractor of the student network mainly follows the structure of the former part of deeplabV3 \cite{dv3}(before ASPP), and use group convolution to replace its original convolutional layers, as shown in TABLE IV. The representation head adopts the structure of stacked $Conv+BN+ReLU$ for feature representation, which is divided into two phases, and adopts bilinear upsampling between phases, as shown in TABLE V.

  \begin{table}[h]
 \centering
\begin{tabular}{c|c|c|l}

\hline
\multicolumn{1}{l|}{} & \multicolumn{1}{l|}{Resolution} & \multicolumn{1}{l|}{Channels} & \multicolumn{1}{c}{Operator} \\ \hline
Stage1                & $\frac{1}{4}$                               & 64                            & $Conv+BN+ReLU+MaxPool$      \\ \hline
Stage2                & $\frac{1}{8}$                               & 256                           & $Bottleneck\times3$              \\ \hline
Stage3                & $\frac{1}{8}$                               & 512                           & $Bottleneck\times4$              \\ \hline
Stage4                & $\frac{1}{8}$                               & 1024                          & $Bottleneck\times6$              \\ \hline
Stage5                & $\frac{1}{8}$                               & 2048                          & $Bottleneck\times3$              \\ \hline
\end{tabular}
\setlength{\abovecaptionskip}{15pt}%
\setlength{\belowcaptionskip}{10pt}%
\caption{Details of the architecture of the feature extractor. The first stage adopts large-stride convolution with max pooling to down-sample the activation map, with expansion of channels between each stage. Bottleneck is proposed by \cite{cf1}. Moreover, the network architecture adopts group convolution \cite{alex} for all convolution layers, and the number of groups is always set to be the greatest common factor of the number of input and output channels.}
\end{table}

\begin{table}[h]
 \centering
\begin{tabular}{c|c|c|l}
\hline
\multicolumn{1}{l|}{} & \multicolumn{1}{l|}{Resolution} & \multicolumn{1}{l|}{Channels} & \multicolumn{1}{c}{Operator}   \\ \hline
Stage1                & $\frac{1}{4}$                               & 512                           & \begin{tabular}[c]{@{}l@{}}$Bilinear+$\\ $(Conv2d+BN+ReLU)\times3$\end{tabular} \\ \hline
Stage2                & $\frac{1}{2}$                               & 512                           & \begin{tabular}[c]{@{}l@{}}$Bilinear+$\\ $(Conv2d+BN+ReLU)\times4$\end{tabular} \\ \hline
Stage3                & 1                               & 2                             & $Bilinear$                    \\ \hline
\end{tabular}
\setlength{\abovecaptionskip}{15pt}%
\setlength{\belowcaptionskip}{10pt}%
\caption{Details of the architecture of the student head. By stacking the bilinear upsampling module with $Conv+BN+ReLU$ formulating as a decoder, resolution expansion and pixel classification are able to realize.}
\end{table}

 \section{PRESENTATION OF DATASET SAMPLES}
 We display all the images in Cityscapes-64 \cite{cityscapes} in Fig. 9, and choose some of the images from KITTI \cite{kitti} exhibit in Fig. 10. As can be seen that none of the images in Cityscapes-64 contain shadows, while some of the images in KITTI have.




\newpage

\begin{figure*}[!t]
	\centering
	\includegraphics[width=7.2in]{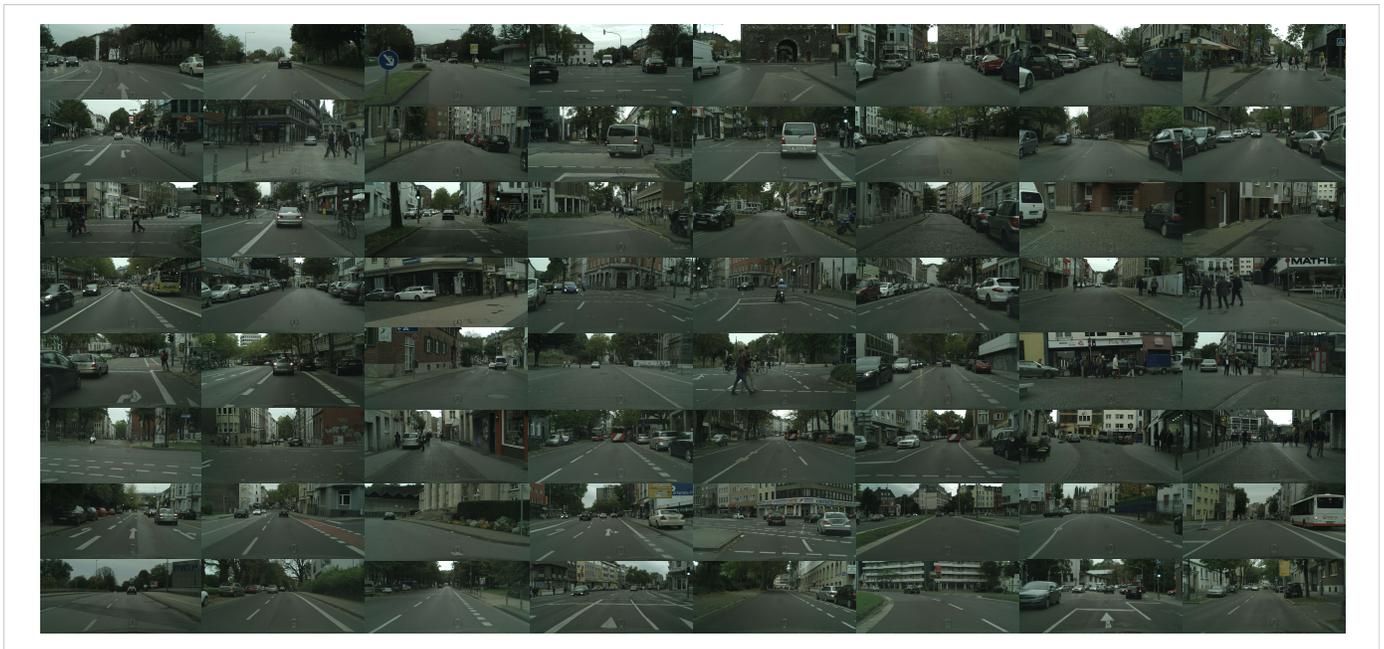}
	\caption{Presentation of all the images in Cityscapes-64 \cite{cityscapes}. All the images do not contain shadows.}
\end{figure*}

\begin{figure*}[!t]
	\centering
	\includegraphics[width=7.2in]{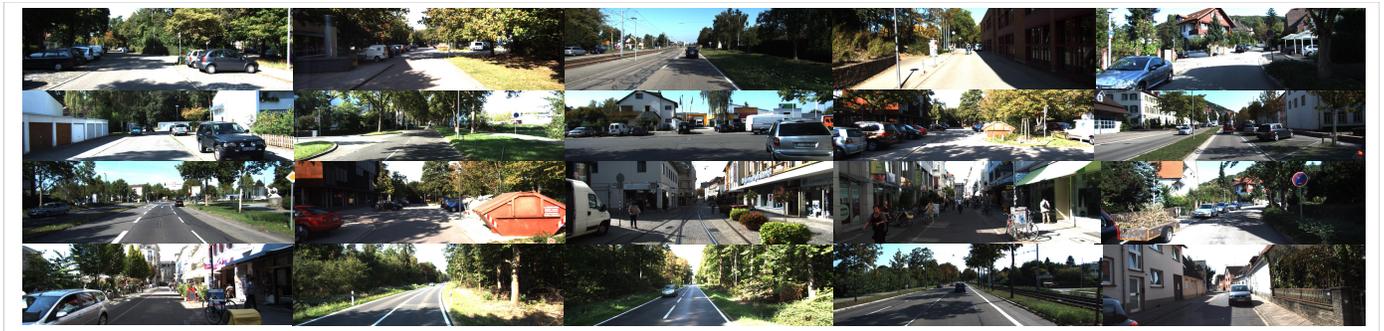}
	\caption{Presentation of some samples in KITTI \cite{kitti}. Some of the images in this dataset contain shadows.}
\end{figure*}

\ifCLASSOPTIONcaptionsoff
  \newpage
\fi




\newpage
\begin{thebibliography}{1}
  
\bibitem{cf1}
He, Kaiming, et al. "Deep residual learning for image recognition." Proceedings of the IEEE conference on computer vision and pattern recognition. 2016.
  
  
  	


\bibitem{app1}
    Grigorescu, Sorin, et al. "A survey of deep learning techniques for autonomous driving." Journal of Field Robotics 37.3 (2020): 362-386.

\bibitem{app2}
    Bojarski, Mariusz, et al. "End to end learning for self-driving cars." arXiv preprint arXiv:1604.07316 (2016).


\bibitem{ss1}
	Sun, Ke, et al. "High-resolution representations for labeling pixels and regions." arXiv preprint arXiv:1904.04514 (2019).

\bibitem{ss2}
Yuan, Yuhui, Xilin Chen, and Jingdong Wang. "Object-Contextual Representations for Semantic Segmentation." arXiv preprint arXiv:1909.11065 (2019).

\bibitem{ss3}
Wang, Li, et al. "Dual Super-Resolution Learning for Semantic Segmentation." Proceedings of the IEEE/CVF Conference on Computer Vision and Pattern Recognition. 2020.




\bibitem{dv3}
Chen, Liang-Chieh, et al. "Rethinking atrous convolution for semantic image segmentation." arXiv preprint arXiv:1706.05587 (2017).

\bibitem{cityscapes}
Cordts, Marius, et al. "The cityscapes dataset for semantic urban scene understanding." Proceedings of the IEEE conference on computer vision and pattern recognition. 2016.

\bibitem{realtime-requirement}
Minaee, Shervin, et al. "Image segmentation using deep learning: A survey." arXiv preprint arXiv:2001.05566 (2020).

\bibitem{rare-situations}
Pouyanfar, Samira, et al. "ROADS: Randomization for obstacle avoidance and driving in simulation." Proceedings of the IEEE/CVF Conference on Computer Vision and Pattern Recognition Workshops. 2019.

\bibitem{da1}
Hariharan, Bharath, and Ross Girshick. "Low-shot visual recognition by shrinking and hallucinating features." Proceedings of the IEEE International Conference on Computer Vision. 2017.

\bibitem{da2}
Chen, Zitian, et al. "Multi-level semantic feature augmentation for one-shot learning." IEEE Transactions on Image Processing 28.9 (2019): 4594-4605.

\bibitem{da3}
Chen, Zitian, et al. "Image block augmentation for one-shot learning." Proceedings of the AAAI Conference on Artificial Intelligence. Vol. 33. No. 01. 2019.

\bibitem{ml1}
Snell, Jake, Kevin Swersky, and Richard S. Zemel. "Prototypical networks for few-shot learning." arXiv preprint arXiv:1703.05175 (2017).

\bibitem{ml2}
Sung, Flood, et al. "Learning to compare: Relation network for few-shot learning." Proceedings of the IEEE conference on computer vision and pattern recognition. 2018.

\bibitem{ml3}
Bertinetto, Luca, et al. "Meta-learning with differentiable closed-form solvers." arXiv preprint arXiv:1805.08136 (2018).

\bibitem{init1}
Finn, Chelsea, Pieter Abbeel, and Sergey Levine. "Model-agnostic meta-learning for fast adaptation of deep networks." International Conference on Machine Learning. PMLR, 2017.

\bibitem{init2}
Rusu, Andrei A., et al. "Meta-learning with latent embedding optimization." arXiv preprint arXiv:1807.05960 (2018).

\bibitem{init3}
Ravi, Sachin, and Hugo Larochelle. "Optimization as a model for few-shot learning." (2016).

\bibitem{finetune}
Yosinski, Jason, et al. "How transferable are features in deep neural networks?." arXiv preprint arXiv:1411.1792 (2014).

\bibitem{liu1}
Liu, Yifan, et al. "Structured knowledge distillation for semantic segmentation." Proceedings of the IEEE Conference on Computer Vision and Pattern Recognition. 2019.

\bibitem{liu2}
Liu, Yifan, et al. "Structured knowledge distillation for dense prediction." IEEE Transactions on Pattern Analysis and Machine Intelligence (2020).

\bibitem{hinton}
Hinton, Geoffrey, Oriol Vinyals, and Jeff Dean. "Distilling the knowledge in a neural network." arXiv preprint arXiv:1503.02531 (2015).

\bibitem{soft-and-hard}
Ba, Jimmy, and Rich Caruana. "Do deep nets really need to be deep?." Advances in neural information processing systems. 2014.

\bibitem{fitnet}
Romero, Adriana, et al. "Fitnets: Hints for thin deep nets." arXiv preprint arXiv:1412.6550 (2014).

\bibitem{he}
He, Tong, et al. "Knowledge adaptation for efficient semantic segmentation." Proceedings of the IEEE Conference on Computer Vision and Pattern Recognition. 2019.

\bibitem{xie}
Xie, Jiafeng, et al. "Improving fast segmentation with teacher-student learning." arXiv preprint arXiv:1810.08476 (2018).


\bibitem{few-survey}
Wang, Yaqing, et al. "Generalizing from a few examples: A survey on few-shot learning." ACM Computing Surveys (CSUR) 53.3 (2020): 1-34.

\bibitem{mobilenetv2}
Sandler, Mark, et al. "Mobilenetv2: Inverted residuals and linear bottlenecks." Proceedings of the IEEE conference on computer vision and pattern recognition. 2018.

\bibitem{mobilenetv3}
Howard, Andrew, et al. "Searching for mobilenetv3." Proceedings of the IEEE International Conference on Computer Vision. 2019.

\bibitem{ghostnet}
Han, Kai, et al. "GhostNet: More features from cheap operations." Proceedings of the IEEE/CVF Conference on Computer Vision and Pattern Recognition. 2020.

\bibitem{cur-learning}
Bengio, Yoshua, et al. "Curriculum learning." Proceedings of the 26th annual international conference on machine learning. 2009.

\bibitem{kd-sum}
Wang, Lin, and Kuk-Jin Yoon. "Knowledge distillation and student-teacher learning for visual intelligence: A review and new outlooks." IEEE Transactions on Pattern Analysis and Machine Intelligence (2021).


\bibitem{alex}
Krizhevsky, Alex, Ilya Sutskever, and Geoffrey E. Hinton. "Imagenet classification with deep convolutional neural networks." Advances in neural information processing systems 25 (2012): 1097-1105.

\bibitem{non-conv}
Denton, Emily, et al. "Exploiting linear structure within convolutional networks for efficient evaluation." arXiv preprint arXiv:1404.0736 (2014).

\bibitem{segnet}
Badrinarayanan, Vijay, Alex Kendall, and Roberto Cipolla. "Segnet: A deep convolutional encoder-decoder architecture for image segmentation." IEEE transactions on pattern analysis and machine intelligence 39.12 (2017): 2481-2495.

\bibitem{kitti}
Geiger, Andreas, Philip Lenz, and Raquel Urtasun. "Are we ready for autonomous driving? the kitti vision benchmark suite." 2012 IEEE Conference on Computer Vision and Pattern Recognition. IEEE, 2012.

\bibitem{coco}
Lin, Tsung-Yi, et al. "Microsoft coco: Common objects in context." European conference on computer vision. Springer, Cham, 2014.


\end{thebibliography}
%


{1}
%

\begin{IEEEbiography}[{\includegraphics[width=1in,height=1.25in,clip,keepaspectratio]{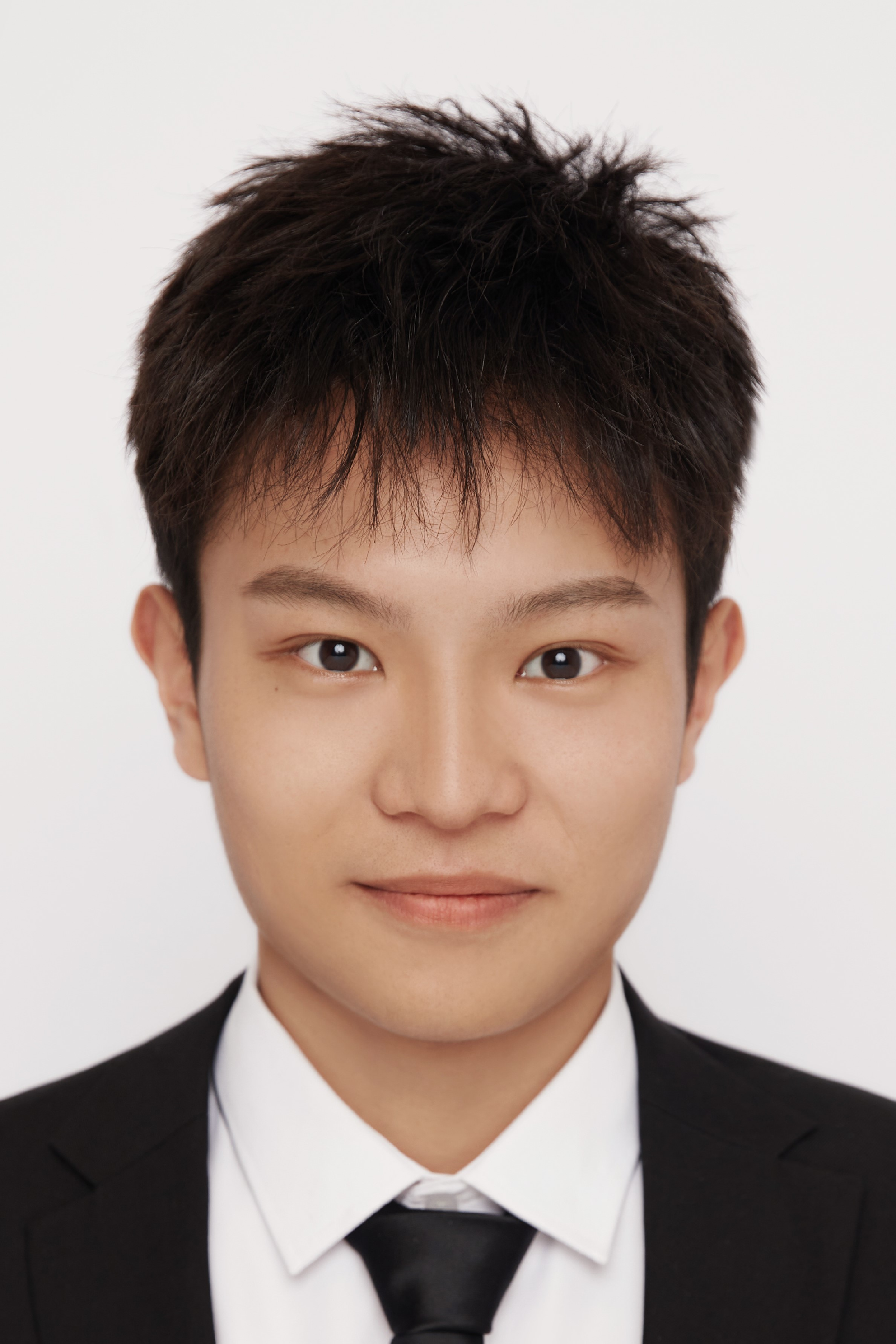}}]{Zhiyuan Wu}
 is currently pursuing the bachelor degree in computer science with the College of Computer Science and Technology, Jilin University, Changchun, China, through the Tang Aoqing Honors Program. His research interests include transfer learning, sensor data fusion and mining, and 3D vision.
\end{IEEEbiography}


\begin{IEEEbiography}[{\includegraphics[width=1in,height=1.25in,clip,keepaspectratio]{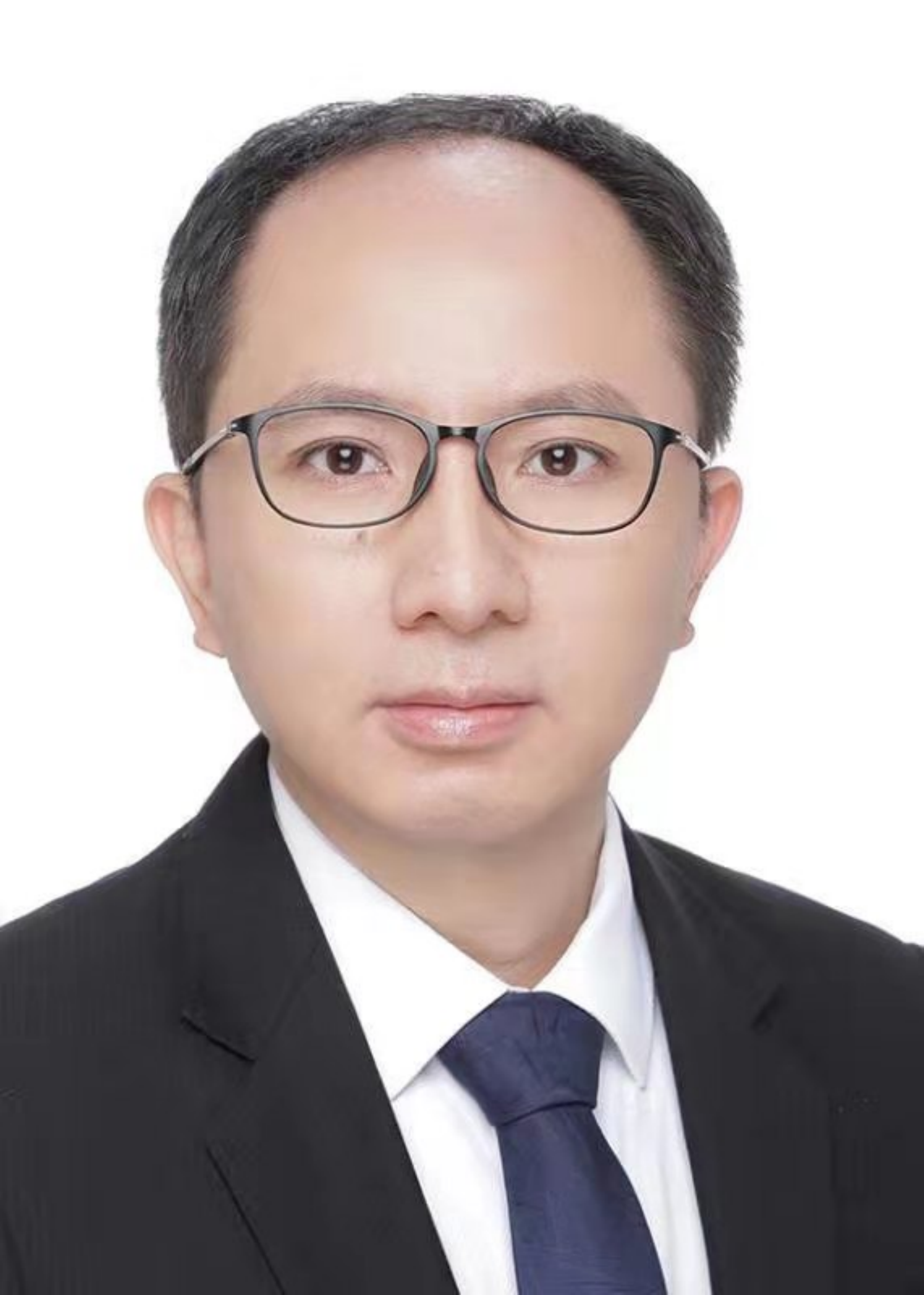}}]{Yu Jiang}
	is an associate professor in the College of Computer Science and Technology at Jilin University. He received his Ph.D. degree from the Jilin University in 2011. His research interests span various topics in sensor data fusion and mining, visual perception, and intelligent control.
\end{IEEEbiography}

\begin{IEEEbiography}[{\includegraphics[width=1in,height=1.25in,clip,keepaspectratio]{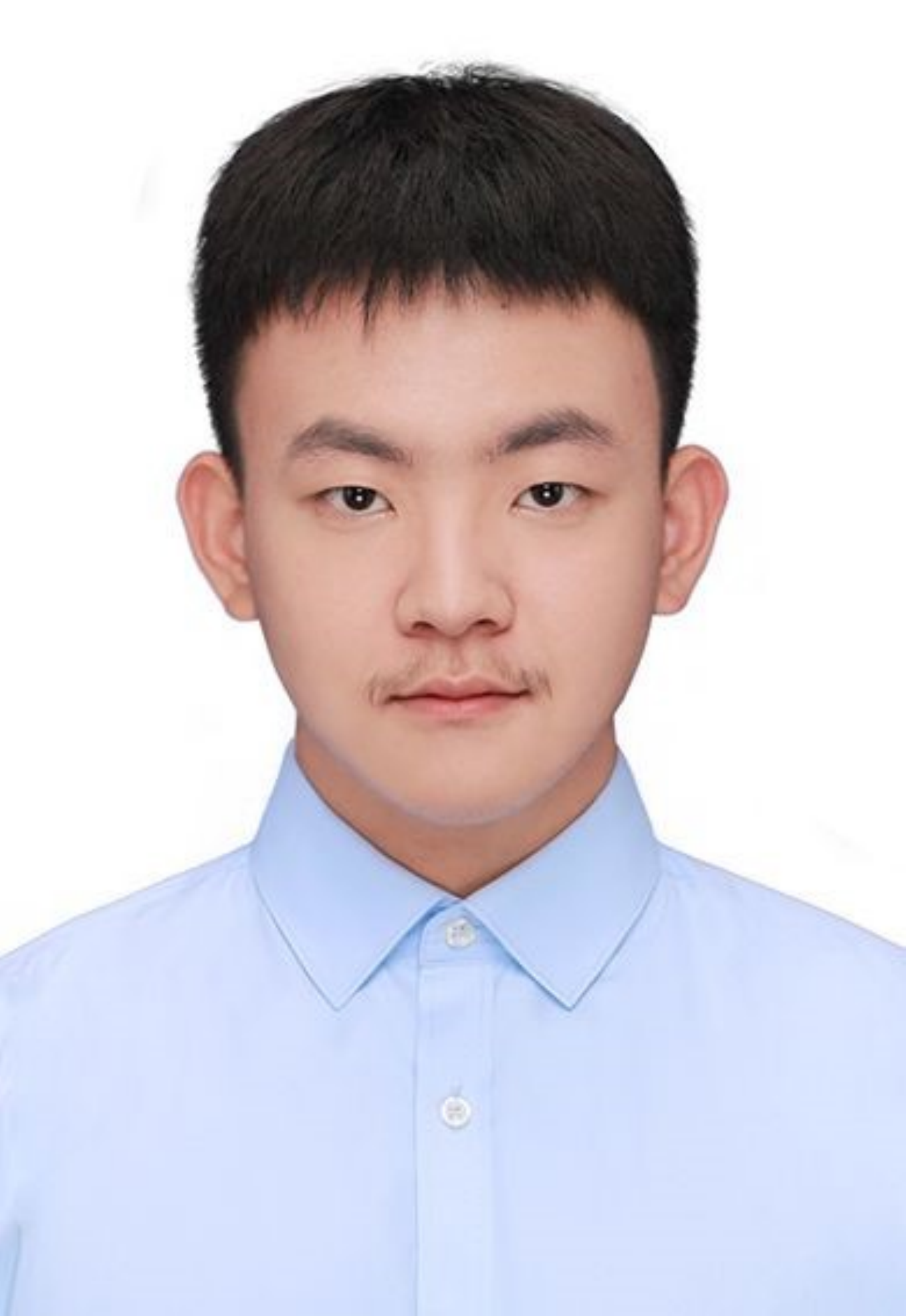}}]{Chupeng Cui}
	is currently pursuing the bachelor degree in computer science with the College of Computer Science and Technology, Jilin University, Changchun, China. His research interests include semantic segmentation and knowledge distillation.
\end{IEEEbiography}

\begin{IEEEbiography}[{\includegraphics[width=1in,height=1.25in,clip,keepaspectratio]{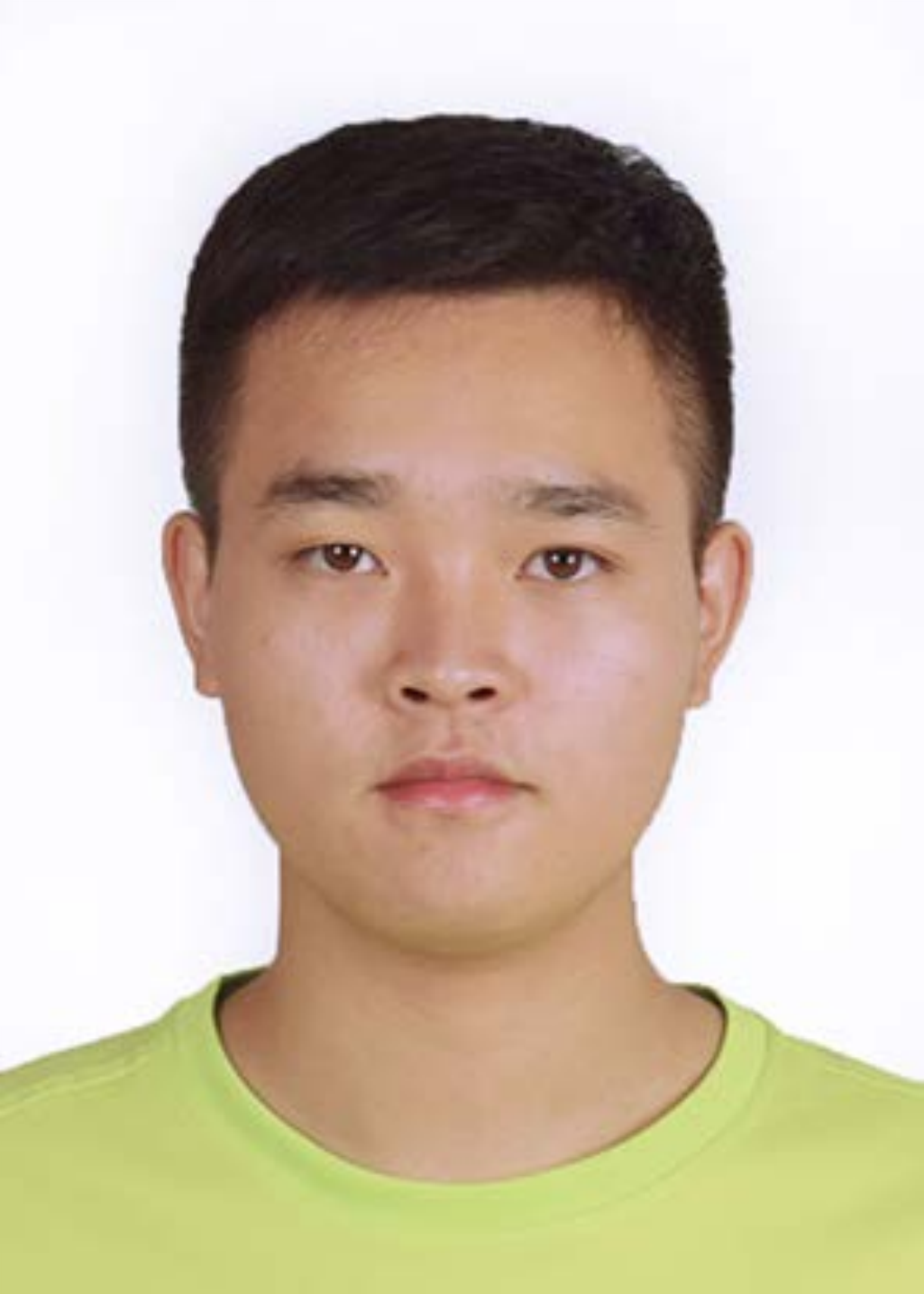}}]{Zongmin Yang}
	is currently pursuing the bachelor degree in computer science with the College of Computer Science and Technology, Jilin University, Changchun, China. His research interests include knowledge distillation and computer vision.
\end{IEEEbiography}

\begin{IEEEbiography}[{\includegraphics[width=1in,height=1.25in,clip,keepaspectratio]{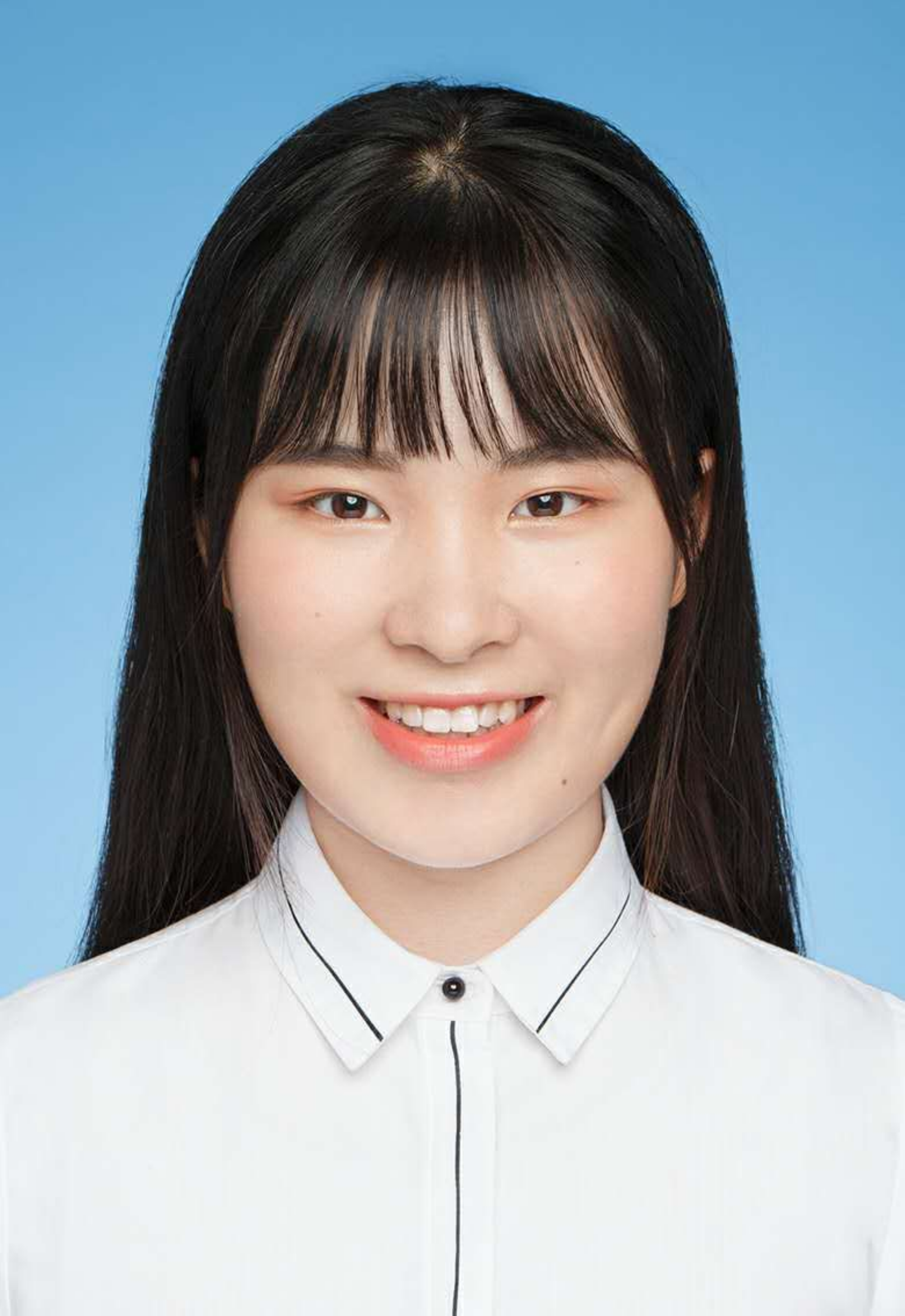}}]{Xinhui Xue}
	is currently pursuing the bachelor degree in computer science with the College of Computer Science and Technology, Jilin University, Changchun, China. Her research interests include transfer learning and anomaly detection.
\end{IEEEbiography}

\begin{IEEEbiography}[{\includegraphics[width=1in,height=1.25in,clip,keepaspectratio]{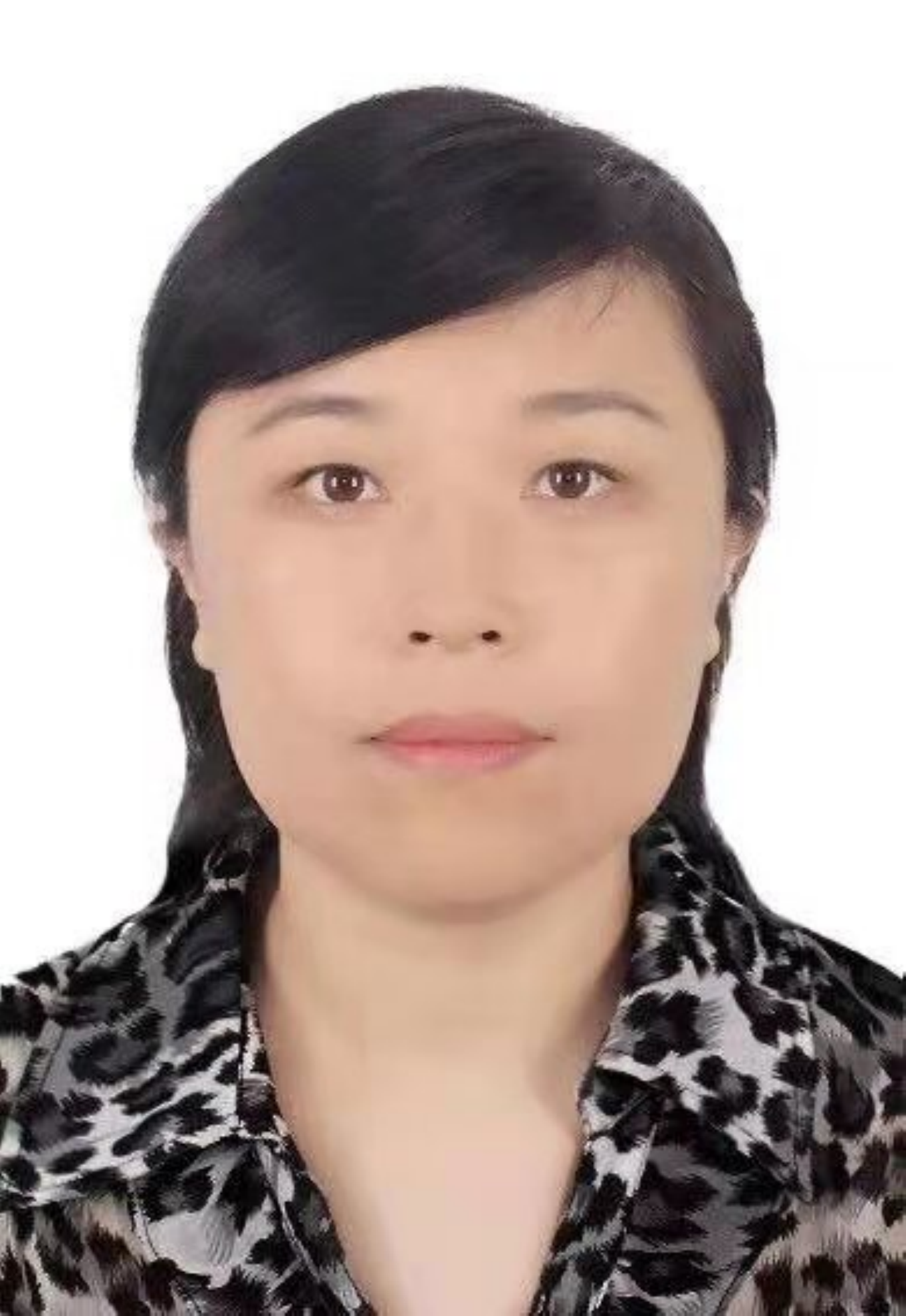}}]{Hong Qi}
 	is an associate professor in the College of Computer Science and Technology, Jilin University. She received the B.S. degree and the Ph.D. degree from Jilin University, Changchun, China. Her research interests span various topics in sensor data fusion and mining, visual perception, and decision making.
\end{IEEEbiography}




\end{document}